\documentclass[12pt,reqno]{amsart}
\usepackage{fullpage}
\usepackage{amsmath}
\usepackage{mathrsfs,amssymb,graphicx,verbatim}
\usepackage{paralist}
\usepackage{amsfonts}
\usepackage{amssymb}
\usepackage{multirow}
\usepackage{euler, latexsym}

\usepackage{ifthen}

\newcommand{\ifsodaelse}[2]{\ifthenelse{\isundefined{\SODAF}}{#2}{#1}}

\ifsodaelse    {\usepackage{ltexpprt}\usepackage{amsmath}}{}
\usepackage{mathrsfs,amssymb,graphicx,verbatim}
\usepackage{paralist}
\usepackage[breaklinks,colorlinks,pdfstartview=FitH]{hyperref}

\newcommand\remove[1]{}

\newcommand{\rnote}[1]{}
\newcommand{\jnote}[1]{}

\date{}

\include{psfig}

\title[Multi-Object Recognition]{Ball-Scale Based Hierarchical Multi-Object Recognition in 3D Medical Images}

\author{Ula\c{s} Ba\u{g}c\i}
\address{School of Computer Science\\ University of Nottingham}
\email{ulasbagc@gmail.com}

\author{Jayaram~K.~Udupa}
\address{Medical Image Processing Group\\UPENN}
\email{jay@mipg.upenn.edu}
\author{ Xinjian Chen}
\address{Diagnostic Radiology\\NIH}
\email{chenx6@mail.nih.gov}
\date{}

\begin{document}
\maketitle

\begin{abstract} This paper investigates, using prior shape models and the concept of ball scale (b-scale), ways of automatically recognizing objects in 3D images without performing elaborate searches or optimization. That is, the goal is to place the model in a single shot close to the right pose (position, orientation, and scale) in a given image so that the model boundaries fall in the close vicinity of object boundaries in the image. This is achieved via the following set of key ideas: (a) A semi-automatic way of constructing a multi-object shape model assembly. (b) A novel strategy of encoding, via b-scale, the pose relationship between objects in the training images and their intensity patterns captured in b-scale images. (c) A hierarchical mechanism of positioning the model, in a one-shot way, in a given image from a knowledge of the learnt pose relationship and the b-scale image of the given image to be segmented. The evaluation results on a set of 20 routine clinical abdominal female and male CT data sets indicate the following: (1) Incorporating a large number of objects improves the recognition accuracy dramatically. (2) The recognition algorithm can be thought as a hierarchical framework such that quick replacement of the model assembly is defined as coarse recognition and delineation itself is known as finest recognition. (3) Scale yields useful information about the relationship between the model assembly and any given image such that the recognition results in a placement of the model close to the actual pose without doing any elaborate searches or optimization. (4) Effective object recognition can make delineation most accurate.
\end{abstract}

\setcounter{tocdepth}{3}
\tableofcontents
\section{Introduction}
The segmentation process as a whole can be thought of as consisting of two tasks: recognition and delineation. Recognition is to determine roughly ``where'' the object is and to distinguish it from other object-like entities. Although delineation is the final step for defining the spatial extent of the object region/boundary in the image, an efficient recognition strategy is a key for successful delineation. In this study, a novel, general method is introduced for object recognition to assist in segmentation (delineation) tasks. It exploits the pose relationship that can be encoded, via the concept of ball scale (b-scale)~\cite{sfc}, between the binary training objects and their associated images.

As an alternative to the manual methods based on initial placement of the models by an expert~\cite{kelemen, pizer} in the literature, model based methods can be employed for recognition. For example, in \cite{soler}, the position of an organ model (such as liver) is estimated by its histogram. In~\cite{brejl}, generalized hough transform is succesfully extended to incorporate variability of shape for 2D segmentation problem. Atlas based methods are also used to define initial position for a shape model. In~\cite{fripp}, affine registration is performed to align the data into an atlas to determine the initial position for a shape model of the knee cartilage. Similarly, a popular particle filtering algorithm is used to detect the starting pose of models for both single and multi-object cases~\cite{bruijne1}. However, due to the large search space and numerous local minimas in most of these studies, conducting a global search on the entire image is not a feasible approach. In this paper, we investigate an approach of automatically recognizing objects in 3D images without performing elaborate searches or optimization. 

\section{Methods}
The proposed method consists of the following key ideas and components:

\textbf{1. Model Building:} After aligning image data from all $N$ subjects in the training set into a common coordinate system via 7-parameter affine registration, the live-wire algorithm~\cite{live_wire} is used to segment $M$ different objects from $N$ subjects. Segmented objects are used for the automatic extraction of landmarks in a slice-by-slice manner~\cite{dryden}. From the landmark information for all objects, a model assembly $MA$ is constructed.

\textbf{2. b-scale encoding:} The b-scale value at every voxel in an image helps to understand
``objectness'' of a given image without doing explicit segmentation. For each voxel, the radius of the largest ball of homogeneous intensity  is weighted by the intensity value of that particular voxel in order to incorporate appearance (texture) information into the object information (called intensity weighted b-scale: $WBs$) so that a model of the correlations between shape and texture can be built. A simple and proper way of thresholding the b-scale image yields a few largest balls remaining in the image. These are used for the construction of the relationship between the segmented training objects and the corresponding images. The resulting images have a strong relationship with the actual delineated objects. 

\textbf{3. Relationship between $MA$ and $WBs$:} A principal component $(PC)$ system is built via PCA for the segmented objects in each image, and their mean $PC$ system, denoted $PC_o$, is found over all training images. $PC_o$ has an origin and three $PC$ axes. Similarly the mean $PC$ system, denoted $PC_b$, for intensity weighted b-scale images $(WBs)$ is found. Finally the transformation $F$ that maps $PC_b$ to $PC_o$ is found. Given an image $I$ to be segmented, the main idea here is to use $F$ to facilitate a quick placement of $MA$ in $I$ with a proper pose as indicated in Step 4 below.

\textbf{4. Hierarchical Recognition:} For a given image $I$,  $WBs$ is obtained and its $PC$ system, denoted $PC_{bI}$ is computed subsequently. Assuming the relationship of $PC_{bI}$ to $PC_o$ to be the same as of $PC_b$ to $PC_o$, and assuming that $PC_o$ offers the proper pose of $MA$ in the training images, we use transformation $F$ and $PC_{bI}$ to determine the pose of $MA$ in $I$. This level of recognition is called coarse recognition. Further refinement of the recognition can be done using the skin boundary object in the image with the requirement that a major portion of $MA$ should lie inside the body region delimited by the skin boundary. Moreover, a little search inside the skin boundary can be done for the fine tuning, however, since offered coarse recognition method gives high recognition rates, there is no need to do any elaborate searches. We will focus on the fine tuning of coarse recognition for future study. The finest level of recognition requires the actual delineation algorithm itself, which is a hybrid method in our case and called GC-ASM (synergistic integration of graph-cut and active shape model). This delineation algorithm is presented in a companion paper submitted to this symposium~\cite{igcasm}.

\section{Model Building}
A convenient way of achieving incorporation of prior information automatically in computing systems is to create and use a flexible \textit{model} to encode information such as the expected \textit{size}, \textit{shape}, \textit{appearance}, and \textit{position} of objects in an image~\cite{davies}. Among such information, \textit{shape} and \textit{appearance} are two complementary but closely related attributes of biological structures in images, and hence they are often used to create statistical models. In particular, shape has been used both in high and low level image analysis tasks extensively, and it has been demonstrated that shape models (such as Active Shape Models (ASMs)) can be quite powerful in compensating for misleading information due to noise, poor resolution, clutter, and occlusion in the images~\cite{hybrid_chen, cremers_kernel, kokkinos}. Therefore, we use ASM~\cite{cootes_asm} to estimate population statistics from a set of examples (training set). In order to guarantee 3D point correspondences required by ASM, we build our statistical shape models combining semi-automatic methods: (1) manually selected anatomically correspondent slices by an expert, and (2) semi-automatic way of specifying key points on the shapes starting from the same anatomical locations. Once Step (1) is accomplished, the remaining problem turns into a problem of establishing point correspondence in 2D shapes, which is easily solved.

\subsection{Establishing Correspondence Across Shapes}
It is extremely significant to choose correct correspondences so that a good representation of the modelled object results. Although landmark correspondence is usually established manually by experts, it is time-consuming, prone to errors, and restricted to only 2D objects~\cite{davies, kendal}. Because of these limitations, a semi-automatic landmark tagging method, \textit{equal space landmarking}, is used to establish correspondence between landmarks of each sample shape in our experiments. Although this method is proposed for 2D objects, and equally spacing a fixed number of points for 3D objects is much more difficult, we use equal space landmarking technique in pseudo-3D manner where 3D object is annotated slice by slice. 

\subsection{Specifying Landmarks}
Let $S \in \mathbb{R}^3$ be a single shape and assume that its finite dimensional representation after the landmarking consisting of $n$ landmark points with positions $\mathbf{LM}^{(i)} \in S, \textit{ }i=1,\ldots, n$, where $\mathbf{LM}^{(i)}=(x^{(i)}, y^{(i)}, z^{(i)})$ are Cartesian coordinates of the $i^{th}$ point on the shape $S$. Equal space landmark tagging for points $\mathbf{LM}^{(i)}$ for $i=1,\ldots,n$ on shape boundaries (contours) starts by selecting an initial point on each shape sample in training set and equally space a fixed number of points on each boundary automatically~\cite{davies}. Selecting the starting point has been done manually by annotating the same anatomical point for each shape in the training set. 

\begin{figure}[h]
\begin{center}
   \begin{tabular}{c}
   \includegraphics[height=5 cm]{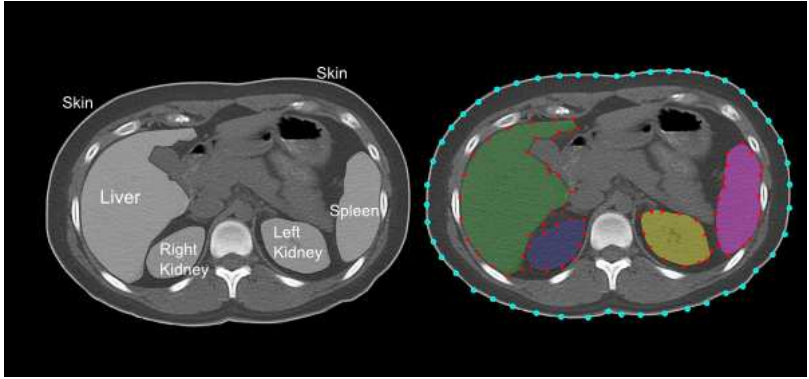}
   \end{tabular}
\caption{A CT slice of abdominal region with selected objects (skin, liver, spleen, and left and right kidney) is shown on the left. Annotated landmarks for the selected objects are shown on the right. \label{img:landmarking_abd}} 
\end{center}
\end{figure}

Figure~\ref{img:landmarking_abd} shows annotated landmarks for five different objects (skin, liver, right kidney, left kidney, spleen) in a CT slice of the abdominal region. Note that different number of landmarks are used for different objects considering their size.

\subsection{Single and Multiple-Object 3D Statistical Shape Models}
\label{sec:asm}
Statistical models of shape variability, or active shape models (ASMs), represent objects as finite number of landmarks and examine the statistics of their coordinates over a number of training sets. The characteristics pattern of a shape class is described by the average shape vector (mean shape) and a linear combination of eigenvectors of the variations around the mean shape, which is formed by simply averaging over all shape samples. An instance of shape from the same object class can then be generated by deforming the mean shape by a linear combination of the retained eigenvectors. For summary of ASM, see~\cite{cootes_asm, chen_oasm}.
In multiple-object ASM, a model assembly $(MA)$ is built, wherein each object class brings  its unique ASM model into the framework. Therefore,  $MA$ can be expressed as a set of models of the form:
\begin{equation}
MA=\left\lbrace \mathbb{M}_1,\ldots, \mathbb{M}_M\right\rbrace, 
\end{equation}
where $M$ denotes the number of objects considered in the model assembly and each model $\mathbb{M}_i=(\mathbf{\overline{x}_i}, \Lambda_i)$ consists of a mean shape $\overline{x}_i$ and allowable variations given by the covariance matrix $\Lambda_i$ for $O_i$, $1 \leq i \leq M$.

\begin{figure}[h]
 \begin{center}
   \begin{tabular}{c}
  \includegraphics[height=4 cm]{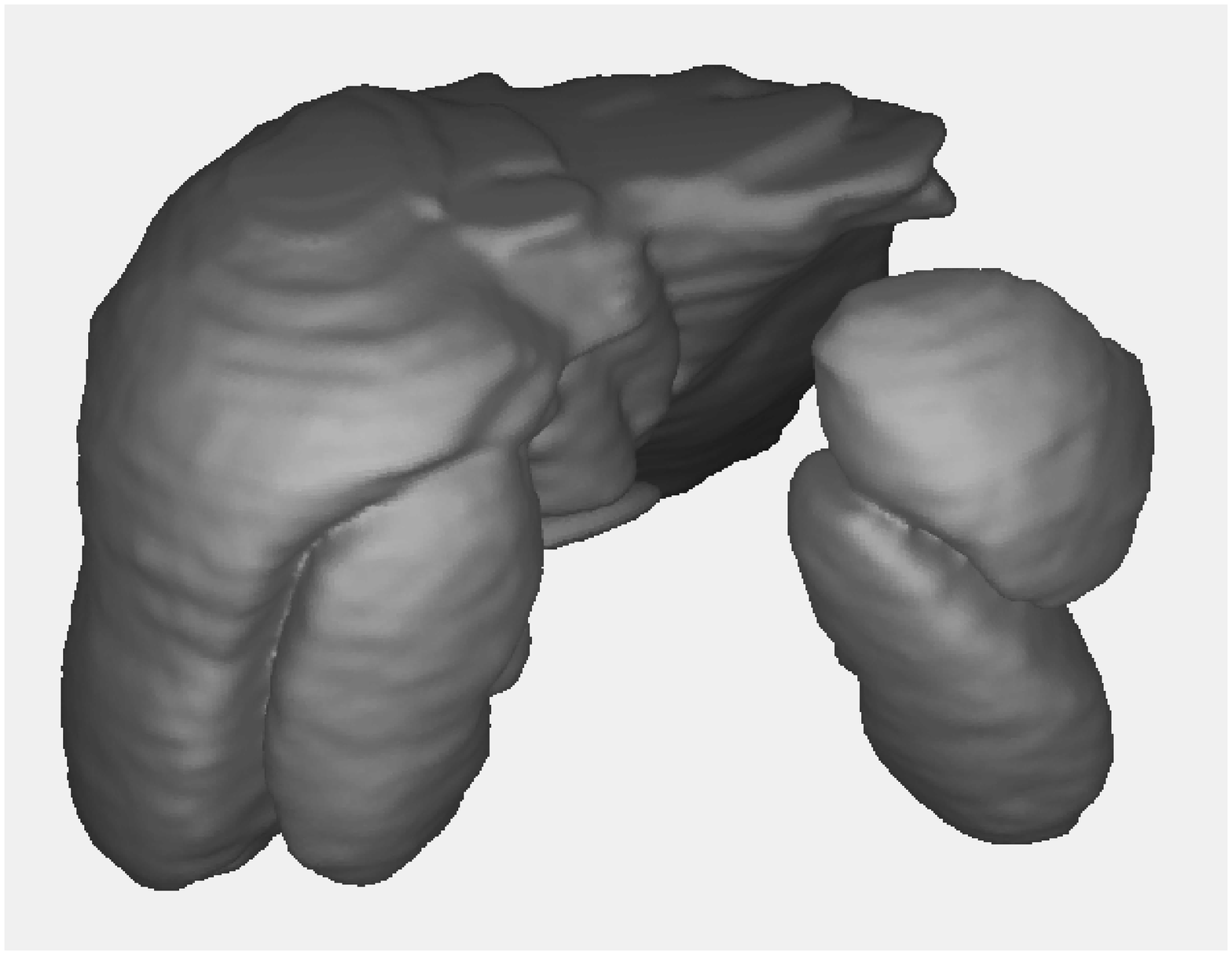}
  \includegraphics[height=4 cm]{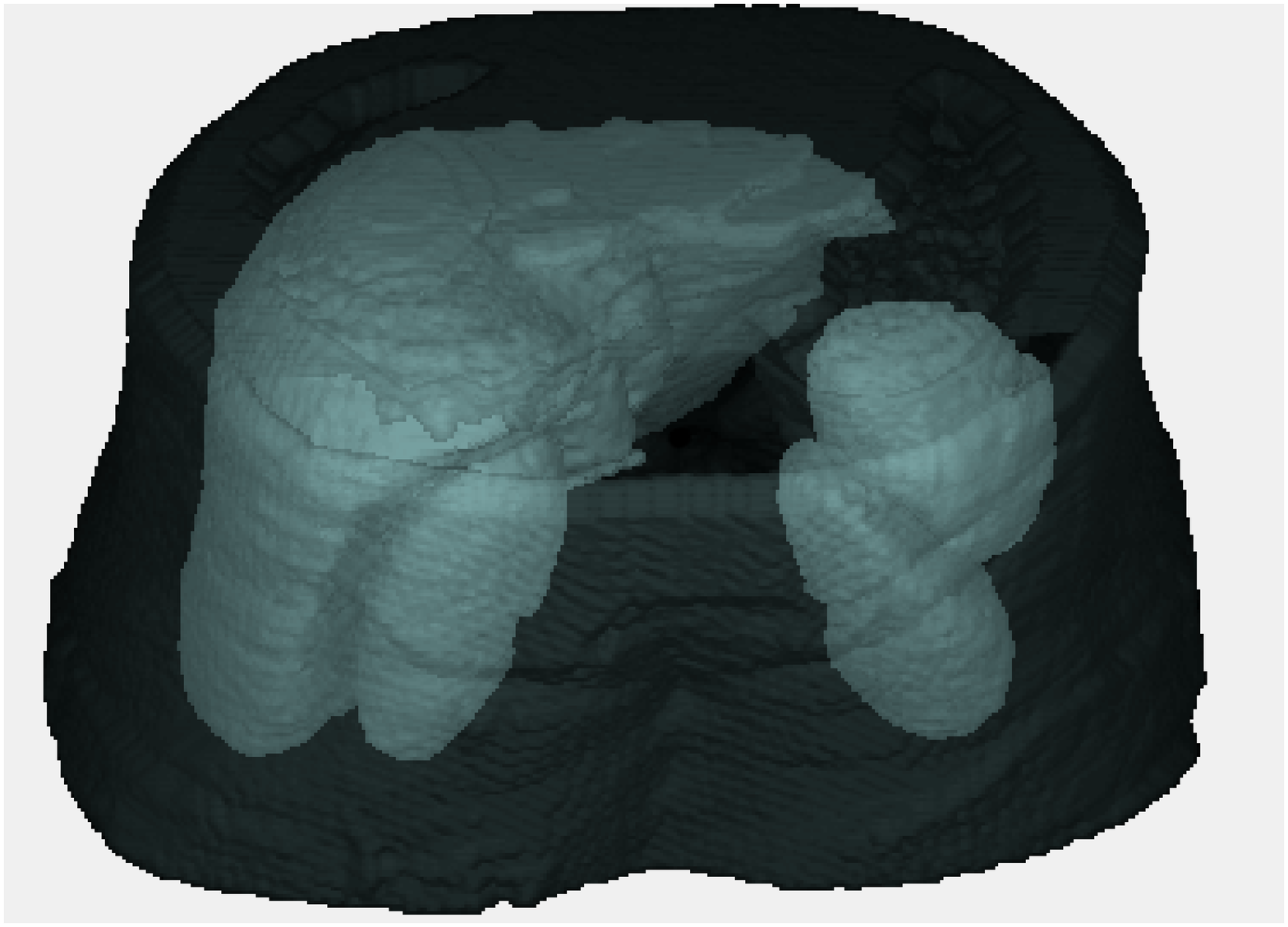}
\end{tabular}
 \end{center}
\caption{Mean shape is generated using 3D ASM for multiple objects of abdominal region. Left: Figure includes mean shapes of  liver, spleen, right and left kidneys.  Right: Figure includes mean shape of skin boundary of the abdominal region as well.\label{img:3dasm_abdominal}} 
\end{figure}

Figure~\ref{img:3dasm_abdominal} shows multi-object 3D ASMs for abdominal organs. Note that skins are also considered in the $MA$s to restrict the search space. Note also that mean shapes of the objects do not have any overlapped region with other mean shapes of the objects. Because, in training part, we select the objects $O_i$ such that $(O_i\cap O_j)_{(i\neq j)\in 1,\ldots, M}=\emptyset$, implying that there is no overlaps across the objects. This fact leads to $(\mathbf{\overline{x}_i}\cap \mathbf{\overline{x}_j})_{(i\neq j)\in 1,\ldots, M}=\emptyset$, as each mean shape $\mathbf{\overline{x}_i}$ is created independently and alignment of the shapes of objects does not affect the distribution of objects in the mean shape due to the nature of the 7-parameter affine transformation $A$:  $(O_i\cap O_j)_{(i\neq j)\in 1,\ldots, M}=\emptyset \Leftrightarrow (A(O_i)\cap A(O_j))_{(i\neq j)\in 1,\ldots, M}=\emptyset \Leftrightarrow (A(\mathbf{\overline{x}_i})\cap A(\mathbf{\overline{x}_j}))_{(i\neq j)\in 1,\ldots, M}=\emptyset$. Objects are not aligned individually, hence, their spatial relations before and after alignment does not change.

Once the $MA$ is created, the next step is to initialize the segmentation process by locating the $MA$ into any given subject image, with proper scale, translation, and orientation parameters, described in the following sections. After the objects in the given image are recognized, local constraints of the shape points in $MA$ are incorporated into the hybrid delineation algorithm called iterative graph-cut active shape model, see \cite{igcasm} for details.

\section{Relationship Between Shape and Intensity Structure Systems}
The goal in recognition of anatomical objects is to identify the extracted shapes of objects. Since extracted geometric patterns are elements of a pattern family which can be thought of as images modulo the invariances represented by the similarity group proposed, they can naturally be considered as desirable image features to roughly identify the relationship of patterns in terms of scale, position, and orientation. Thus, we conjecture that creating a pattern family that includes rough object information together with region information yields concise bases for recognizing objects. For this purpose, without doing explicit segmentation, a rough but definitive representation of objects is possible by the b-scale approach. Hence, we have endeavoured to integrate locally adaptive b-scale information of objects into the recognition process to produce geometric patterns extracted by a b-scale based filtering method using region homogeneity.

There are several advantages to the scale based approach. First, boundary and region based representations of objects are explicitly contained in the scale-based methods. Based on continuity of homogeneous regions inside images, we roughly identify geometric properties of objects, namely scale information, and represent the actual images with this new representation, called scale images, i.e. ball-scale, tensor-scale, generalized-scale images. Second, since scale based methods are able to identify objects embodied in the images roughly, resultant rough objects can be used as prior information to be integrated into the whole segmentation process. Third, and most importantly, scale images provide fast recognition of objects with high accuracy. Scale based methods provide highly accurate estimates of position, orientation, and size of the actual objects such that there is no need to do elaborate searches within the images to locate the objects. To the best of our knowledge, this is the only existing study for 3D images locating objects of interest for a given image in one shot, namely without doing any search. In the following subsection, we describe how shape and intensity structure systems are related through b-scale based method.


\subsection{Ball Scale Encoding}
\label{sec:am}
\begin{figure}[h]
\begin{center}
   \begin{tabular}{c}
   \includegraphics[height=5 cm]{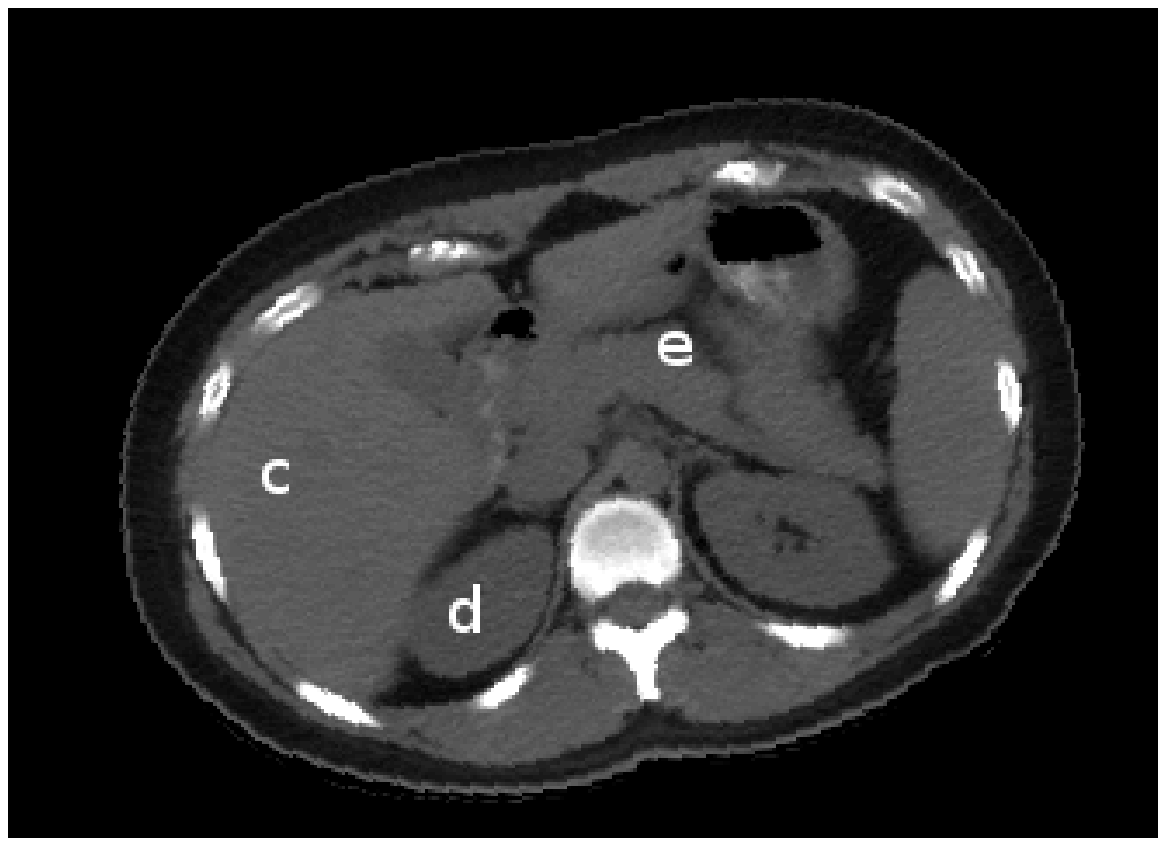}
   \includegraphics[height=5 cm]{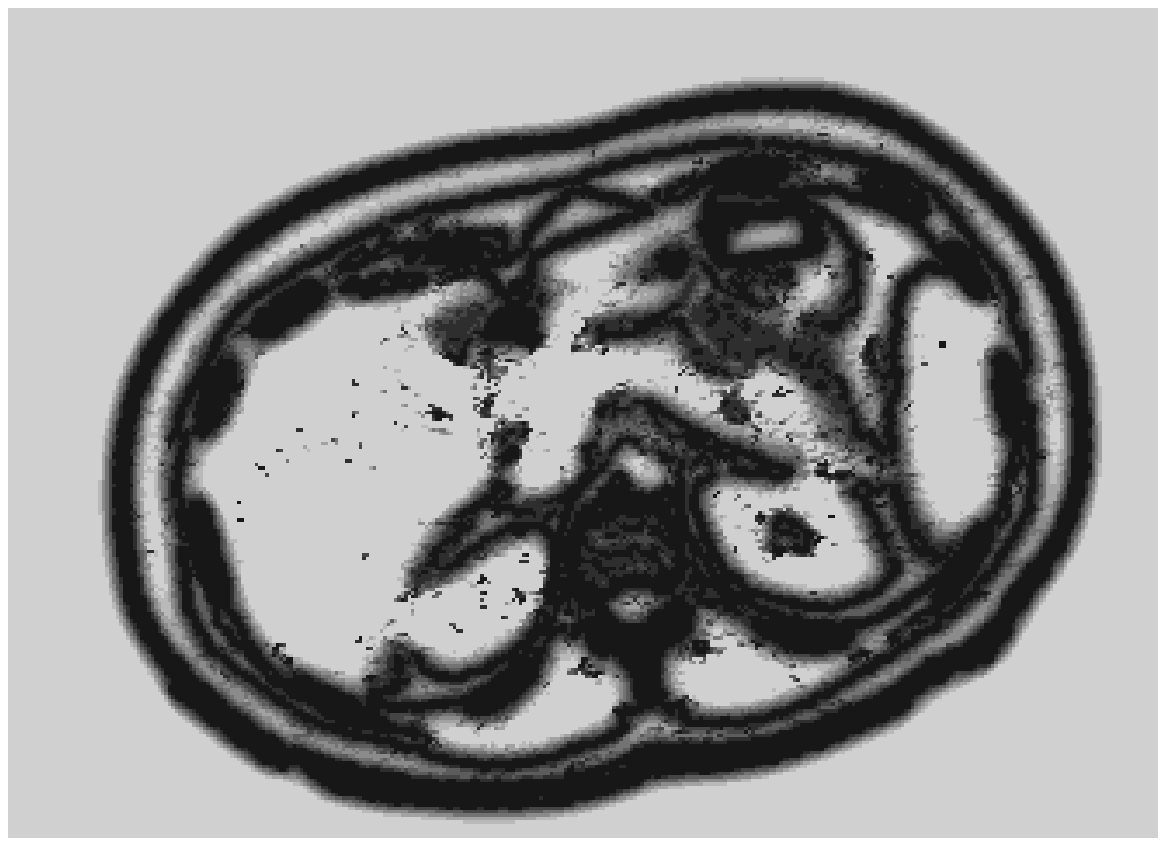} 
   \end{tabular}
   \end{center}
\caption{ (a) A 2D slice from a 3D-CT scene of an abdominal region. Using a suitable region-homogeneity criterion, a local determination is done as to what is the largest disc that can be centered at any point such as $c$ within which the intensities are homogeneous. The b-scale at $c$ is bigger than that $d$ or at $e$. (b) The b-scale scene of the CT abdominal slice in (a). \label{img:bscale}} 
\end{figure}

The b-scale method has been reported to be very useful in image segmentation~\cite{sfc}, filtering~\cite{bscale}, inhomogeneity correction\cite{zhuge_inhomog}, and image registration~\cite{reg_udupa}. The main idea in b-scale encoding is to determine the size of local structures at every spel under a prespecified scene-dependent region-homogeneity criterion. For example, in Figure~\ref{img:bscale}, the size of the hyperball located at pixel $c$ is bigger than that located $d$ or $e$, thus the size of the local structure to which pixel $c$ belongs is bigger than that to which $d$ or  $e$ belongs. By definition, and also seen from the Figure~\ref{img:bscale}, locally adaptive scale in regions with fine details or in the vicinity of boundaries is small, while it is large in the interior of a large homogeneous object regions.

\subsubsection{Computation of b-scale}
\textbf{Terminology \& Notation:} The pair $(\mathbb{Z}^n, \alpha)$ is called a \textit{digital space} where $\alpha$ is an adjacency relation on $\mathbb{Z}^n$. We represent a $nD$ image over a fuzzy digital space $(Z^n, \alpha)$, called \textit{scene} for short, by a pair $\mathcal{C}=(C,f)$ where $C$ is a finite nD array of spels (spatial elements), called \textit{scene domain} of  $\mathcal{C}$, covering a body region of the particular patient for whom image data $\mathcal{C}$ are acquired, and $f$ is an intensity function defined on $C$, which assigns an integer intensity value to each spel $c \in C$. We assume that $f(c) \geq 0$ for all $c \in C$ and $f(c)=0$ if and only if there are no measured data for spel $c$ and that $\nu$ is an n-tuple indicating the physical dimensions of the spels in $C$.

A \textit{hyperball} $B_{k,\nu}$ of radius $k \geq 0$ and with center at $c \in C$ in a scene $\mathcal{C}=(C,f)$ over $(Z^n,\alpha,\nu)$ is defined by
\begin{equation}
B_{k,\nu}=\left\lbrace e \in C  \Biggr \vert  \sqrt{\sum_{i=1}^n\dfrac{\nu_i^2(c_i - e_i)^2}{min_j\left[ \nu_j^2\right] }} \leq k \right\rbrace .
\end{equation}
A fraction $FO_{k,\nu}(c)$ (``fraction of object"), that indicates the fraction of the ball boundary occupied by a region which is sufficiently homogeneous with $c$, by
\begin{equation}
FO_{k,\nu}(c)=\dfrac{\sum_{e \in B_{k,\nu}(c)-B_{k-1,\nu}(c)}W_\psi(|f(c)-f(e)|)}{|B_{k,\nu}(c) - B_{k-1,\nu}(c)|}
\end{equation}
where $|B_{k,\nu}(c)-B_{k-1,\nu}(c)|$ is the number of spels in $B_{k,\nu}(c)-B_{k-1,\nu}(c)$ and $W_{\psi}$ is a homogeneity function~\cite{bscale, mada}. A detailed description of the characteristics of $W_{\psi}$ is presented in~\cite{bscale}. In all experiments, we use a zero-mean unnormalized Gaussian function for $W_{\psi}$.

The ball radius $k$ is iteratively increased starting from one and the b-scale computation algorithm checks for $FO_{k,\nu}(c)$, the fraction of the object containing $c$ that is contained in the ball. When this fraction falls below the threshold $t_s$, it is considered that the ball contains an object region different from that to which $c$ belongs~\cite{bscale}. Following the recommendation in~\cite{sfc}, $t_s=0.85$ is chosen.

\subsubsection{Intensity Weighted Ball Scale - $WB_{scale}$}
\label{subsub:wbs}
Although the size of a local structure is estimated using the appearance information of the gray scale images, i.e. region-homogeneity criterion, b-scale images contain only rough geometric information. Incorporating appearance information into this rough knowledge characterizes scale information of local structures, thus, it allows us to distinguish objects with the same size by their appearance information. One way to extract b-scale images with corresponding appearance information of the local structures is to weight the radius of the hyperball centered at a given spel with the intensity value of that spel. As a result, object scale information is enriched with local intensity values.

The algorithm for intensity weighted object scale estimation (IWOSE) is presented below.\\
\textbf{Algorithm} IWOSE\\
\textbf{Input:} $c \in C$ in a scene $\mathcal{C}=(C,f)$, $W_{\psi}$, a fixed threshold $t_s$\\ 
\textbf{Output:} $r'(c)$\\
1. begin\\
2.\quad set $k=1$\\
3. \quad while $FO_{k,\nu}(c) \geq t_s$ do\\
4. \quad \quad set $k$ to $k+1$\\
5. \quad endwhile;\\
6. \quad set $r(c)$ to $k-1$;\\
7. \quad output $r'(c)=f(c)r(c)$;\\
8. end\\

\begin{figure}[h]
\begin{center}
   \begin{tabular}{c}
   \includegraphics[height=2.5 cm]{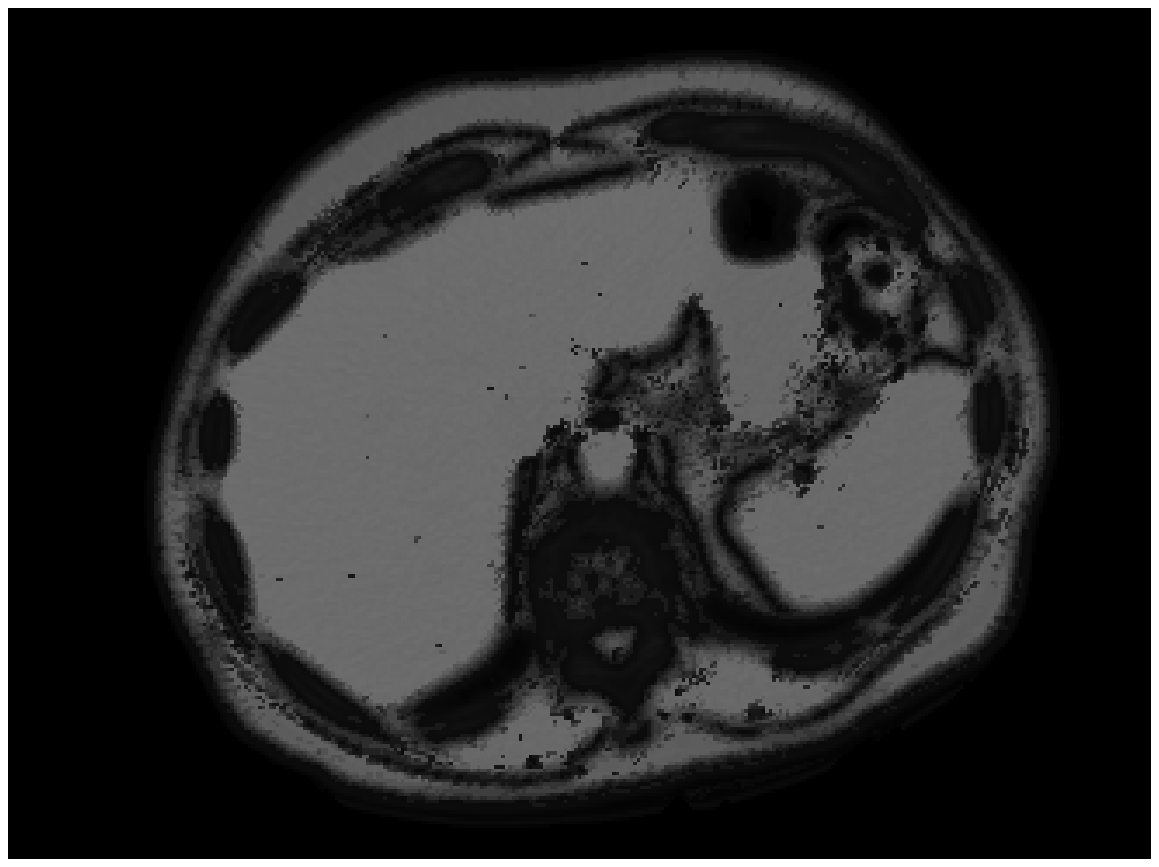}
   \includegraphics[height=2.5 cm]{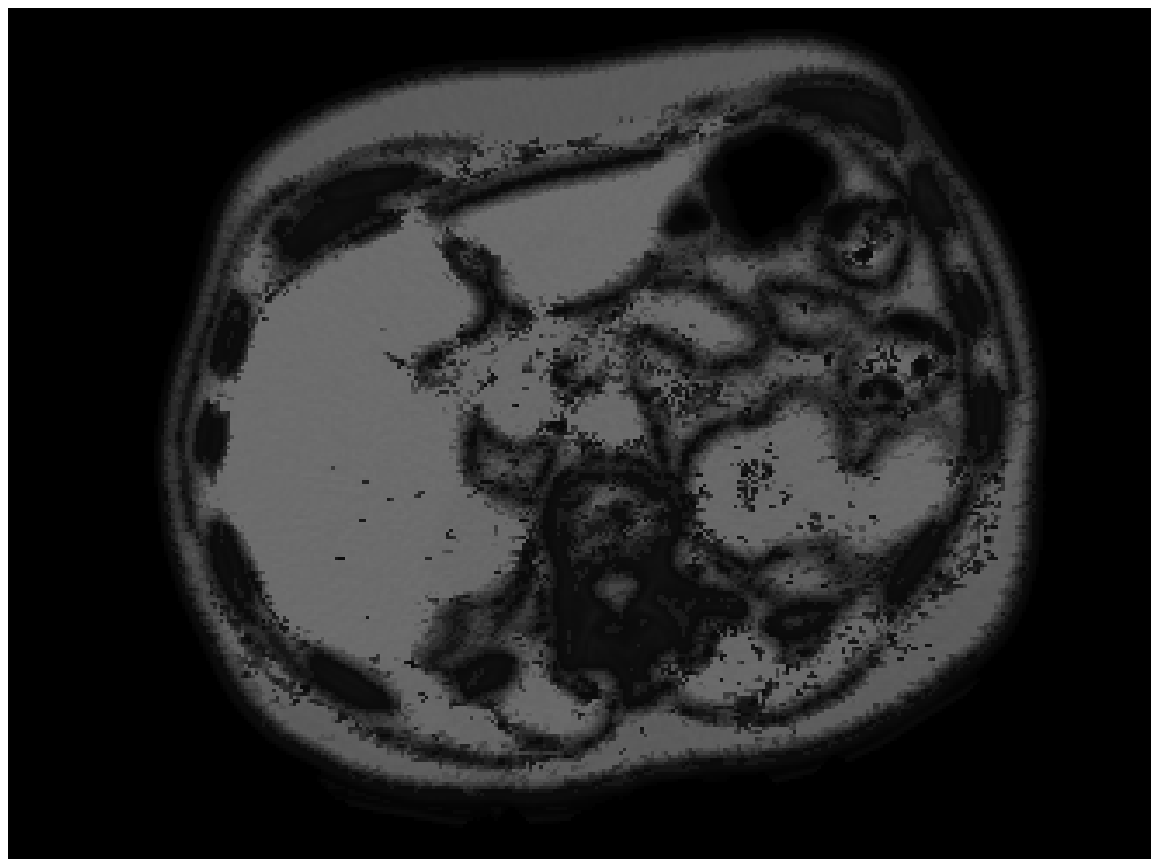}
   \includegraphics[height=2.5 cm]{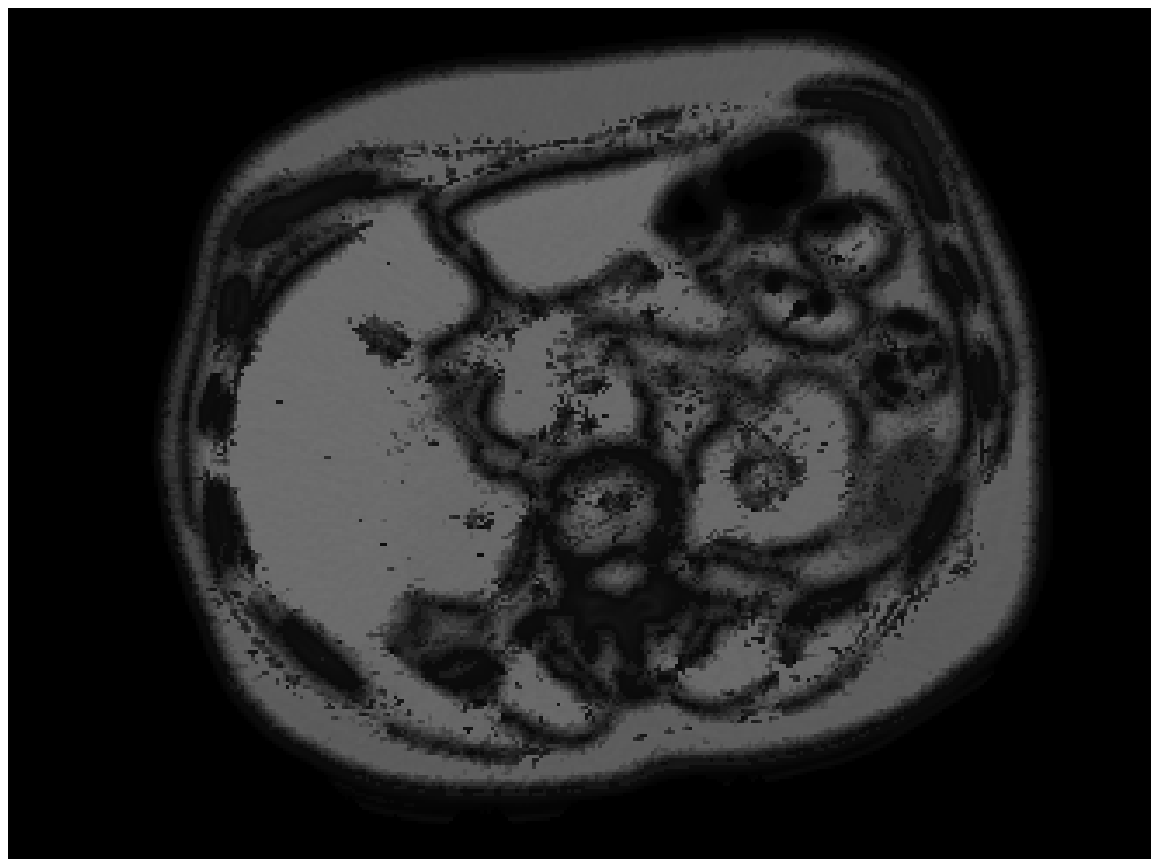}
   \includegraphics[height=2.5 cm]{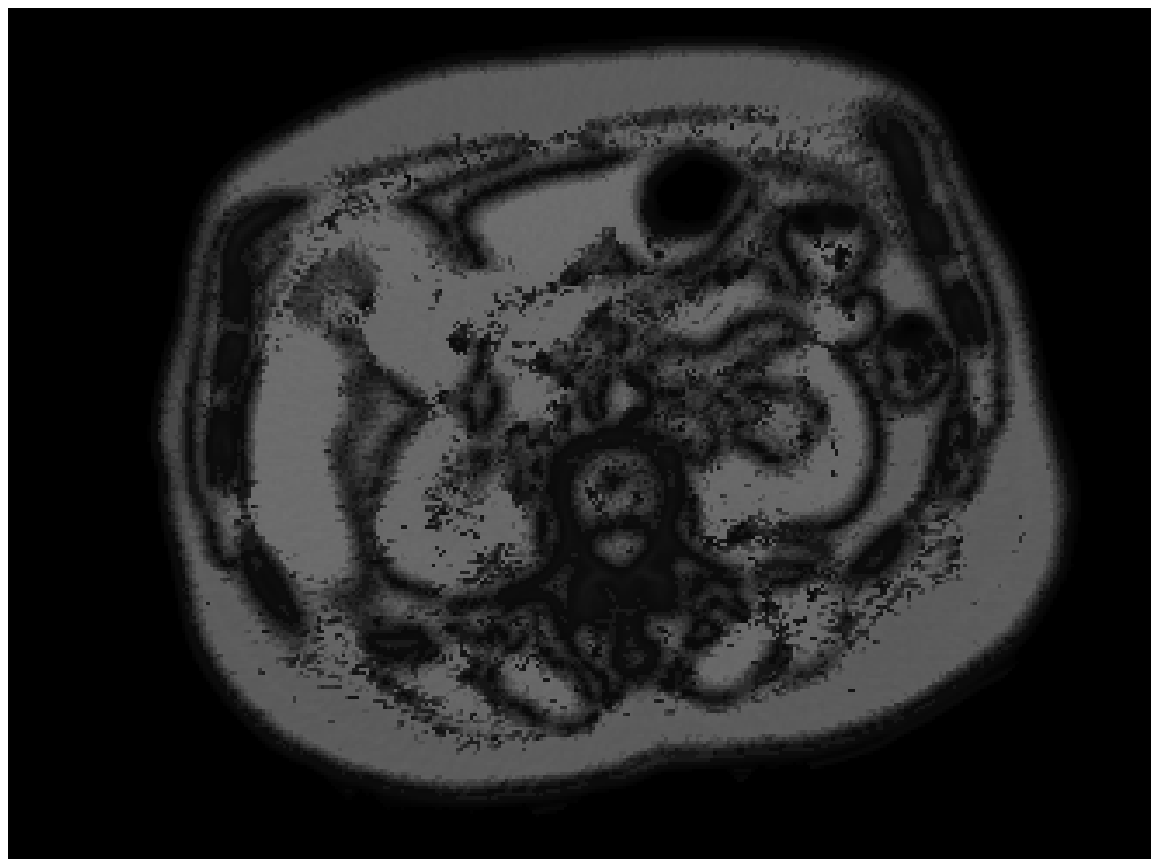}\\
   \includegraphics[height=2.5 cm]{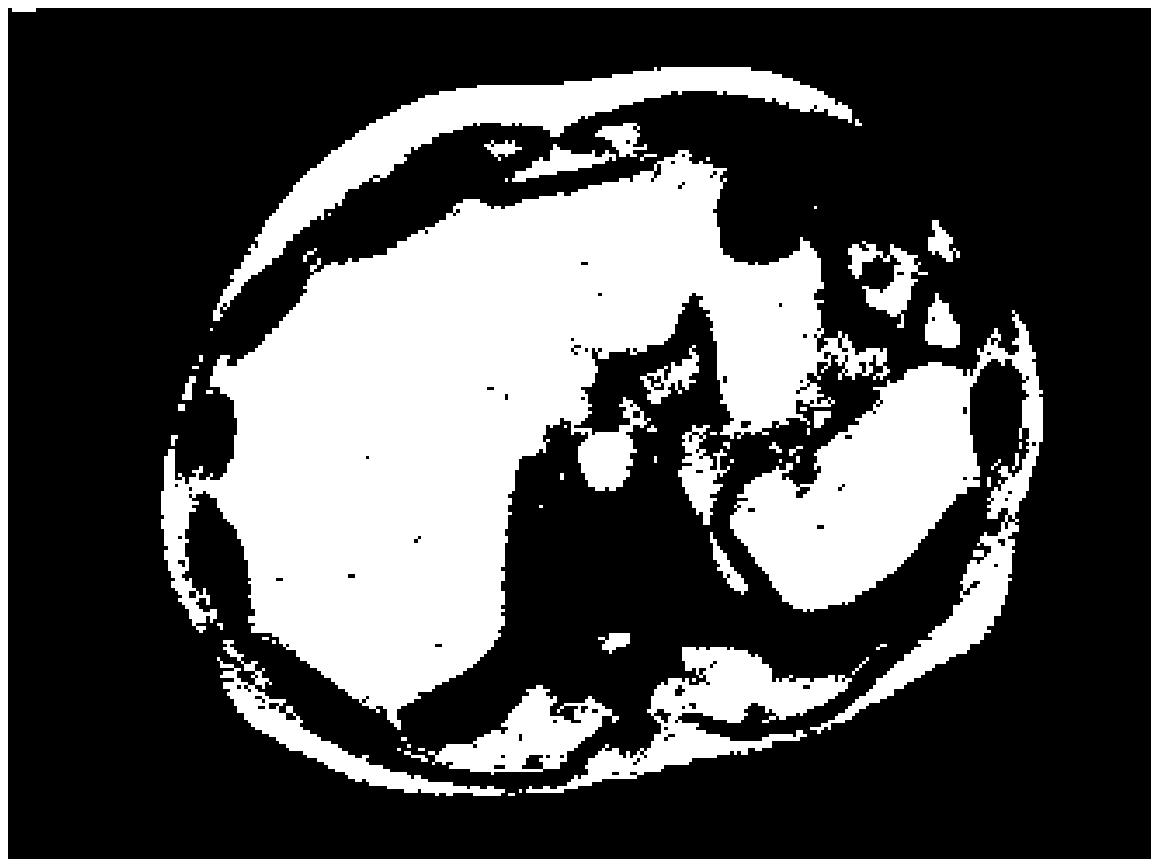}
   \includegraphics[height=2.5 cm]{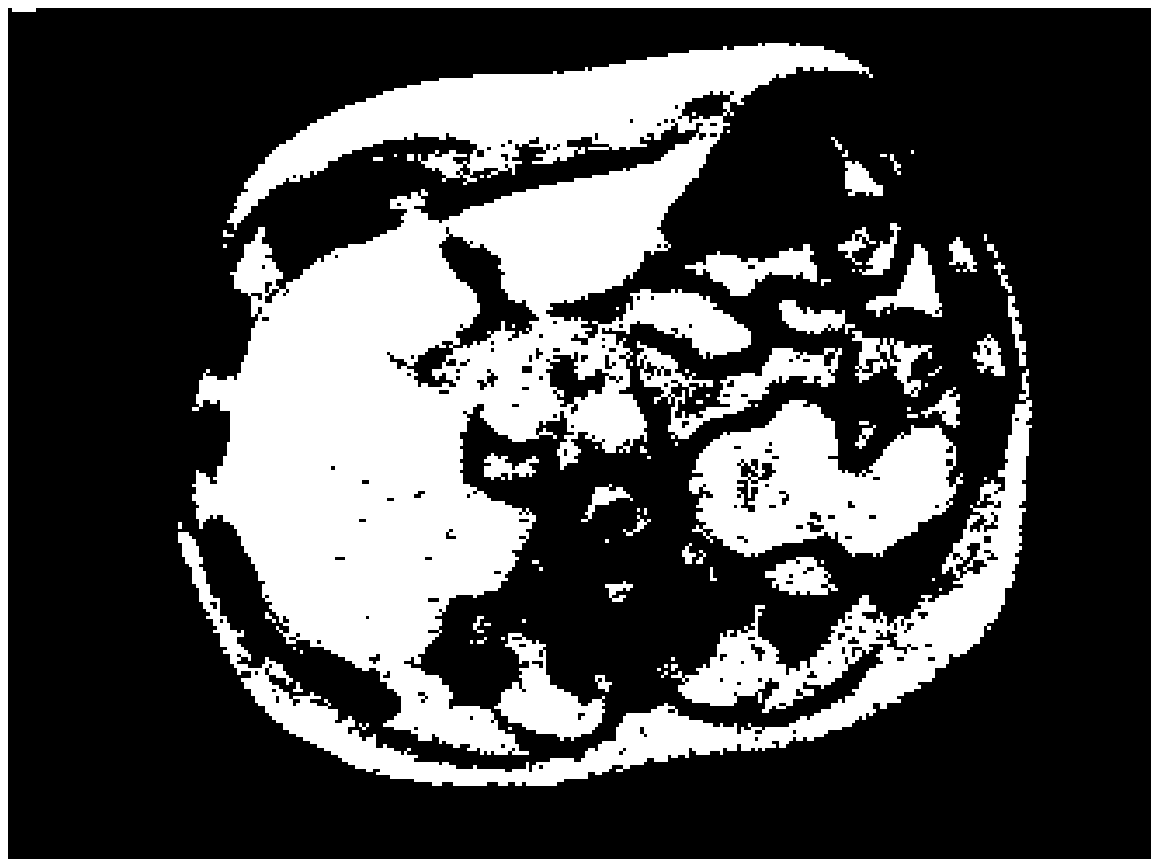}
   \includegraphics[height=2.5 cm]{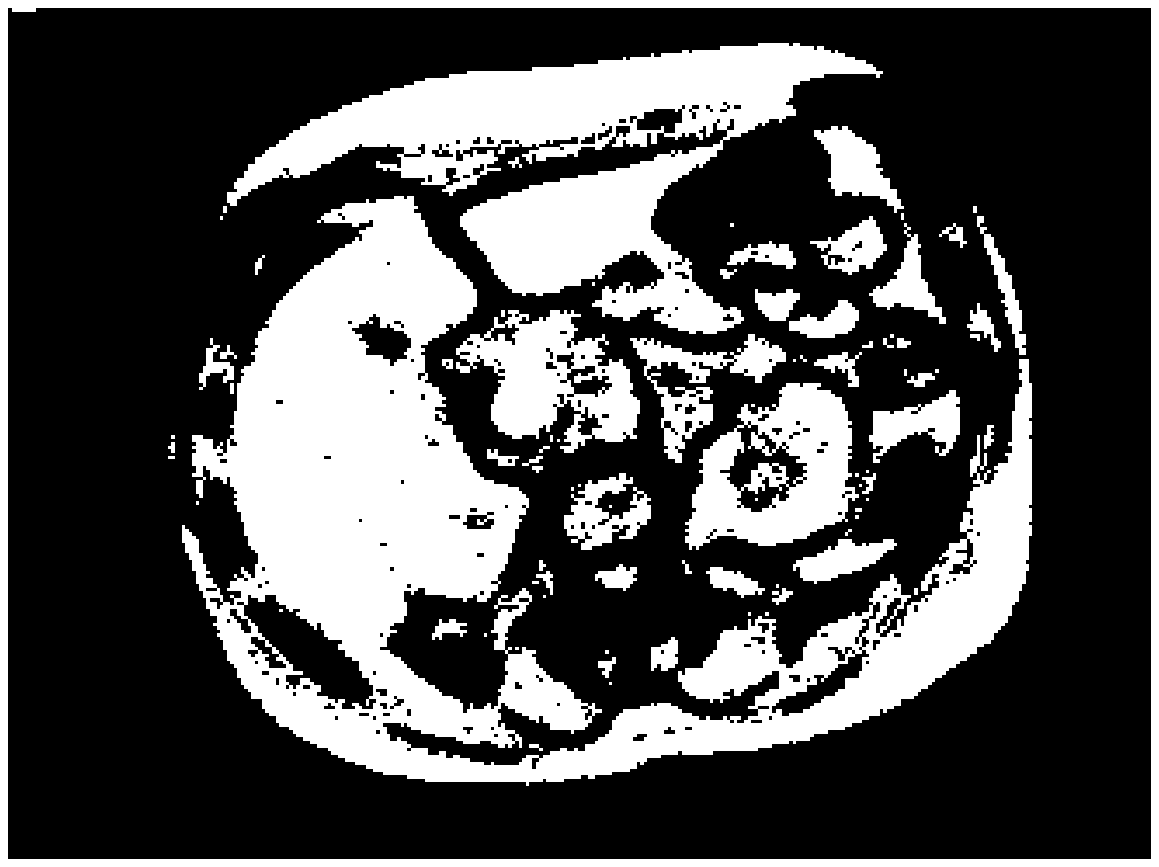}
   \includegraphics[height=2.5 cm]{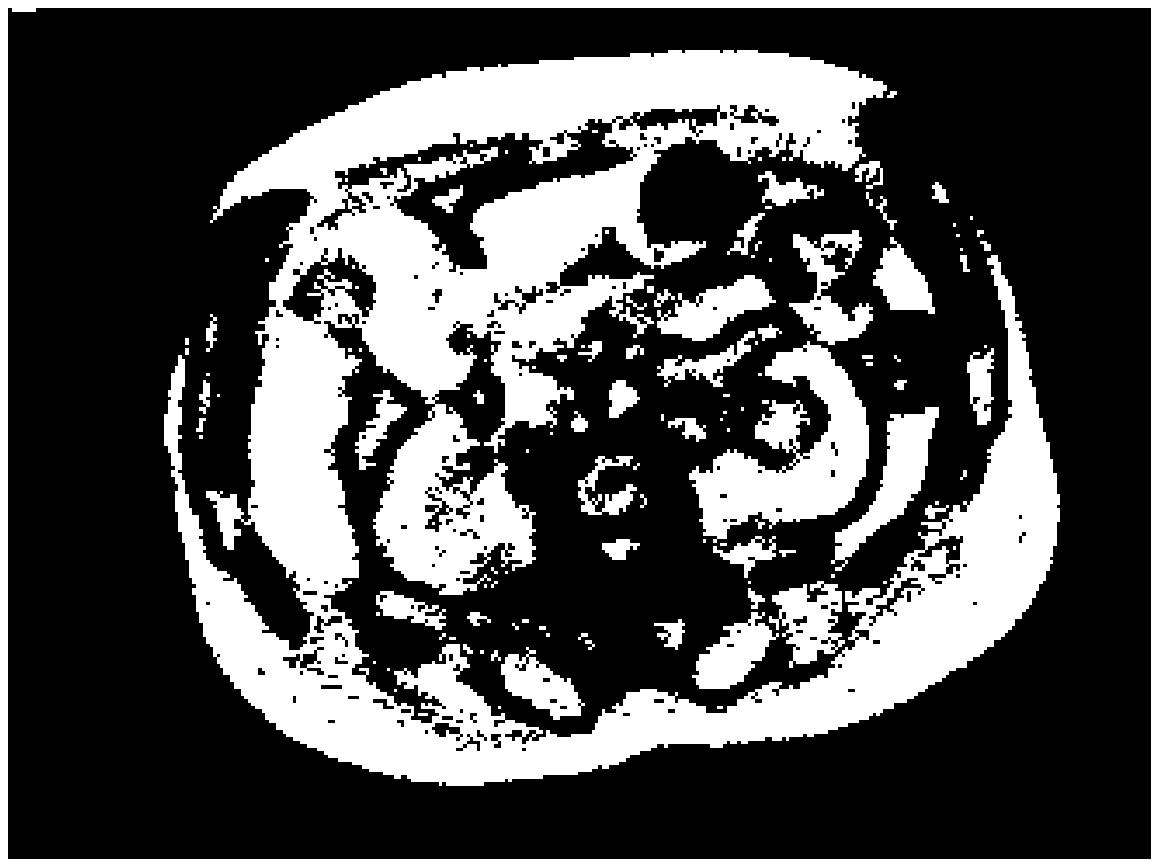}\\ 
      \includegraphics[height=2.5 cm]{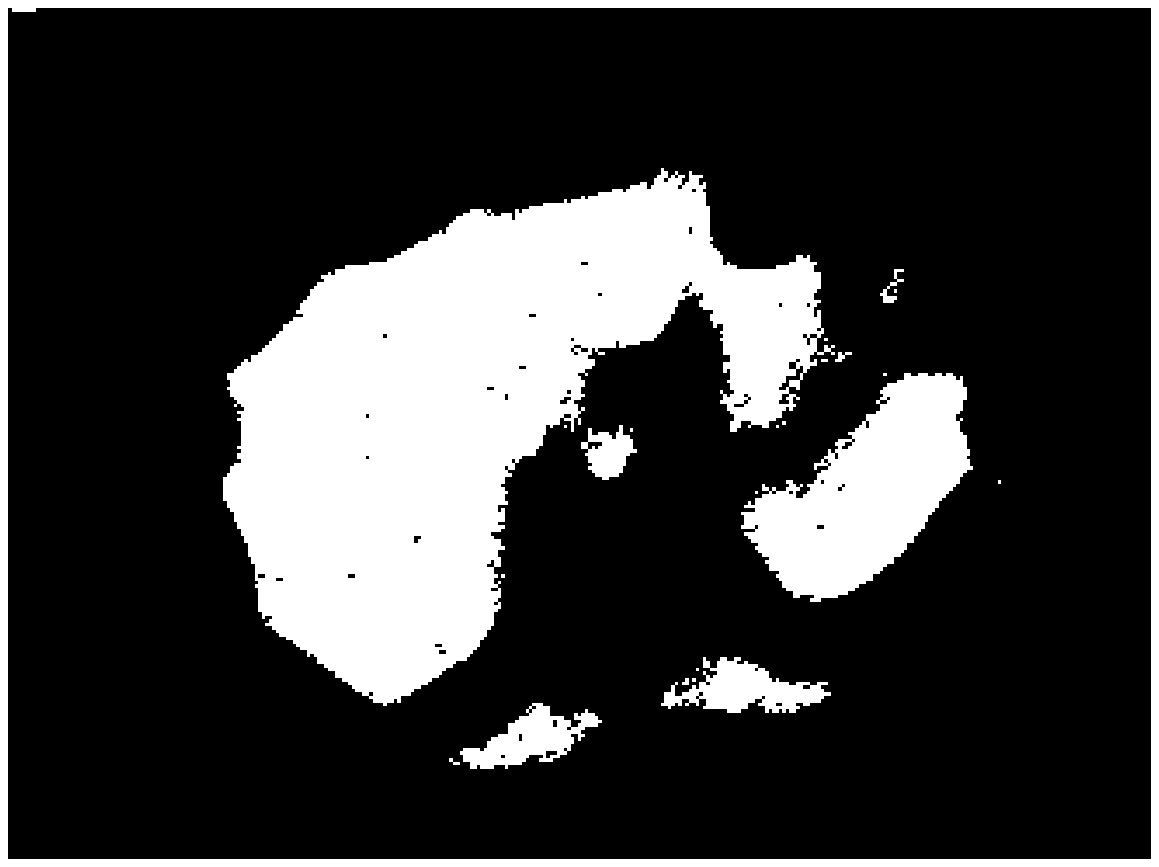}
   \includegraphics[height=2.5 cm]{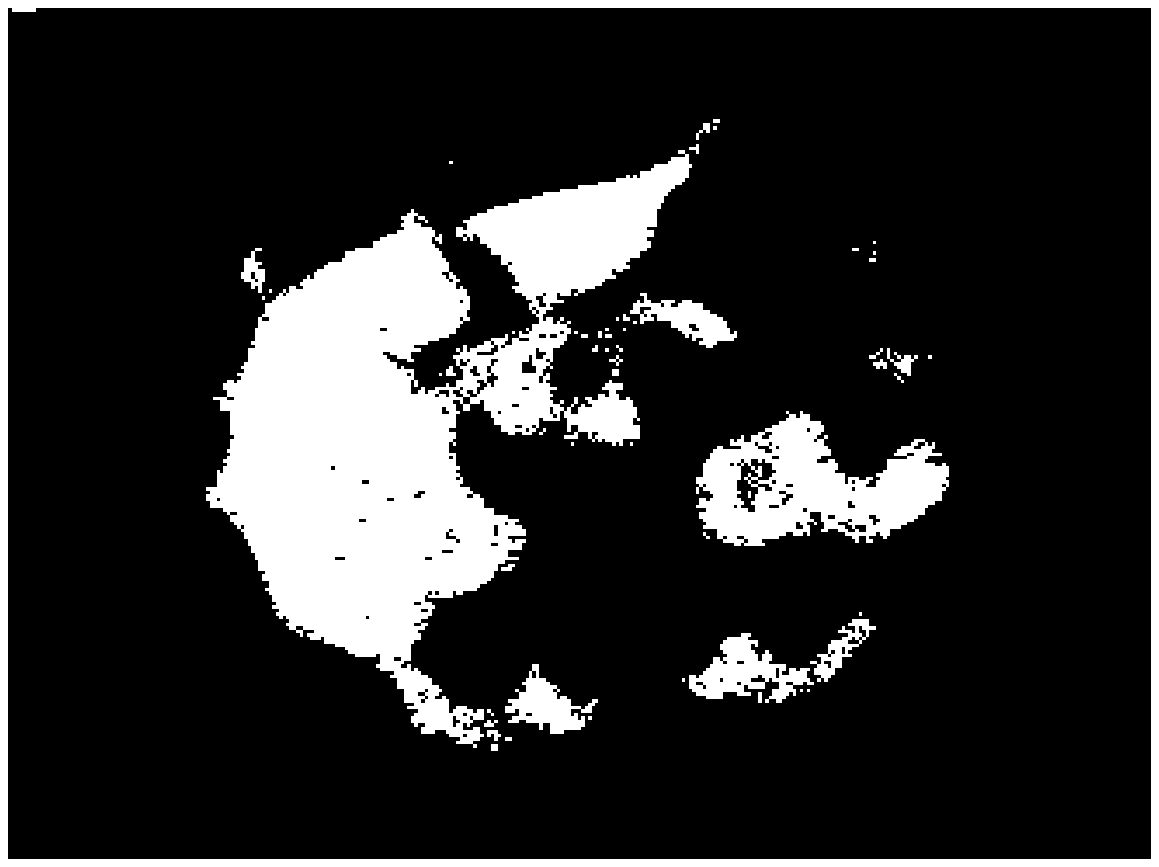}
   \includegraphics[height=2.5 cm]{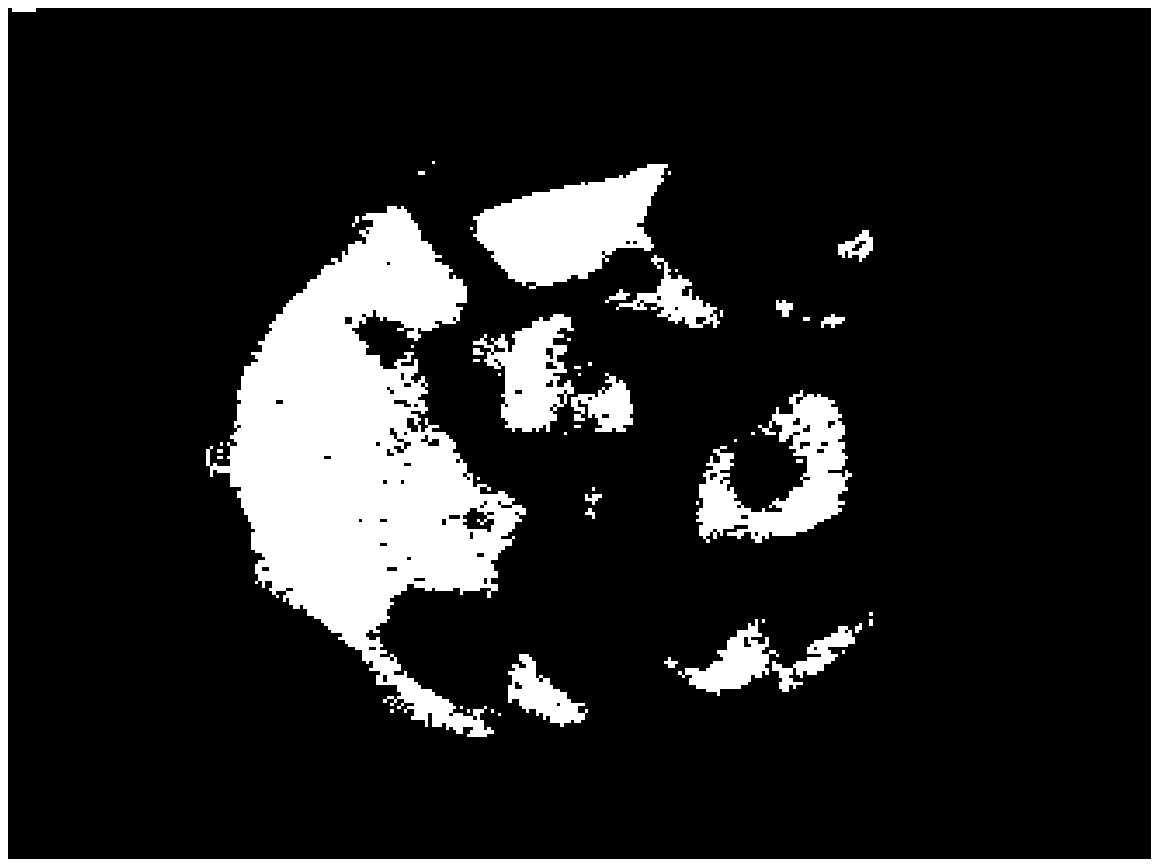}
   \includegraphics[height=2.5 cm]{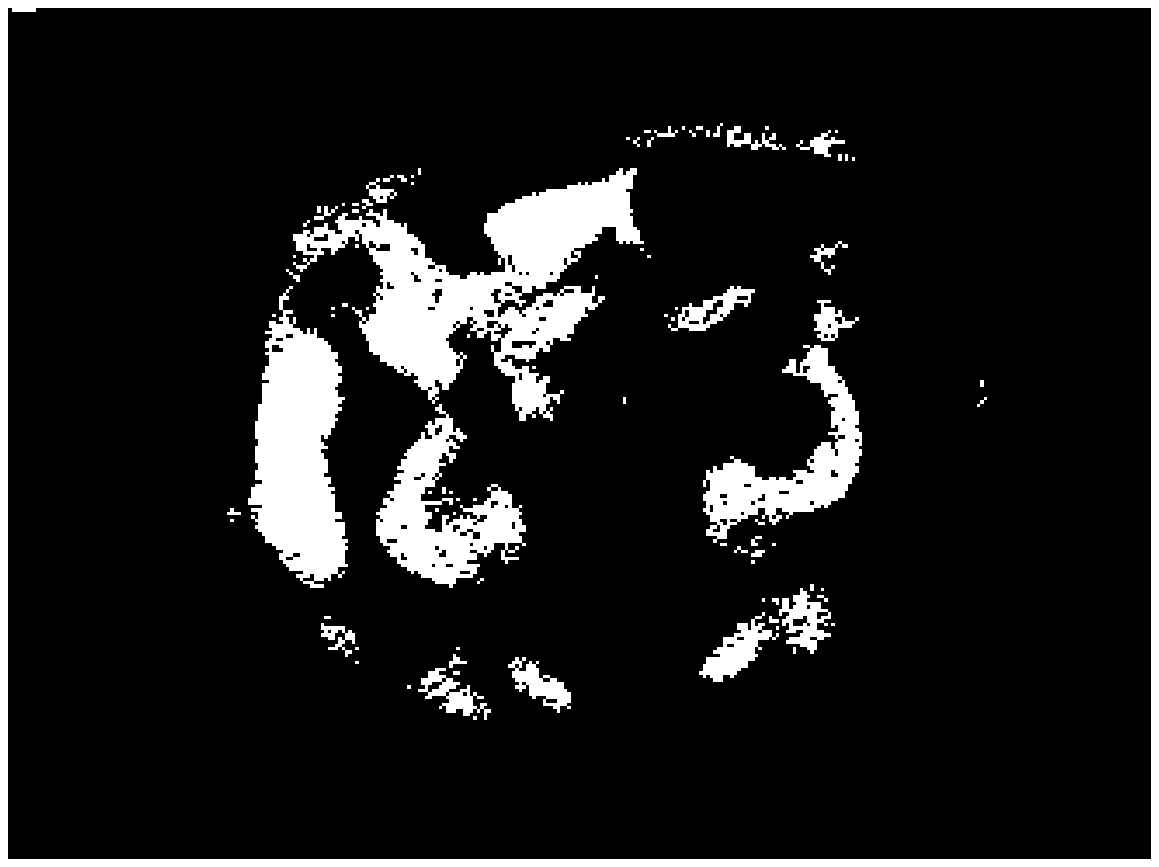}\\ 
      \includegraphics[height=2.5 cm]{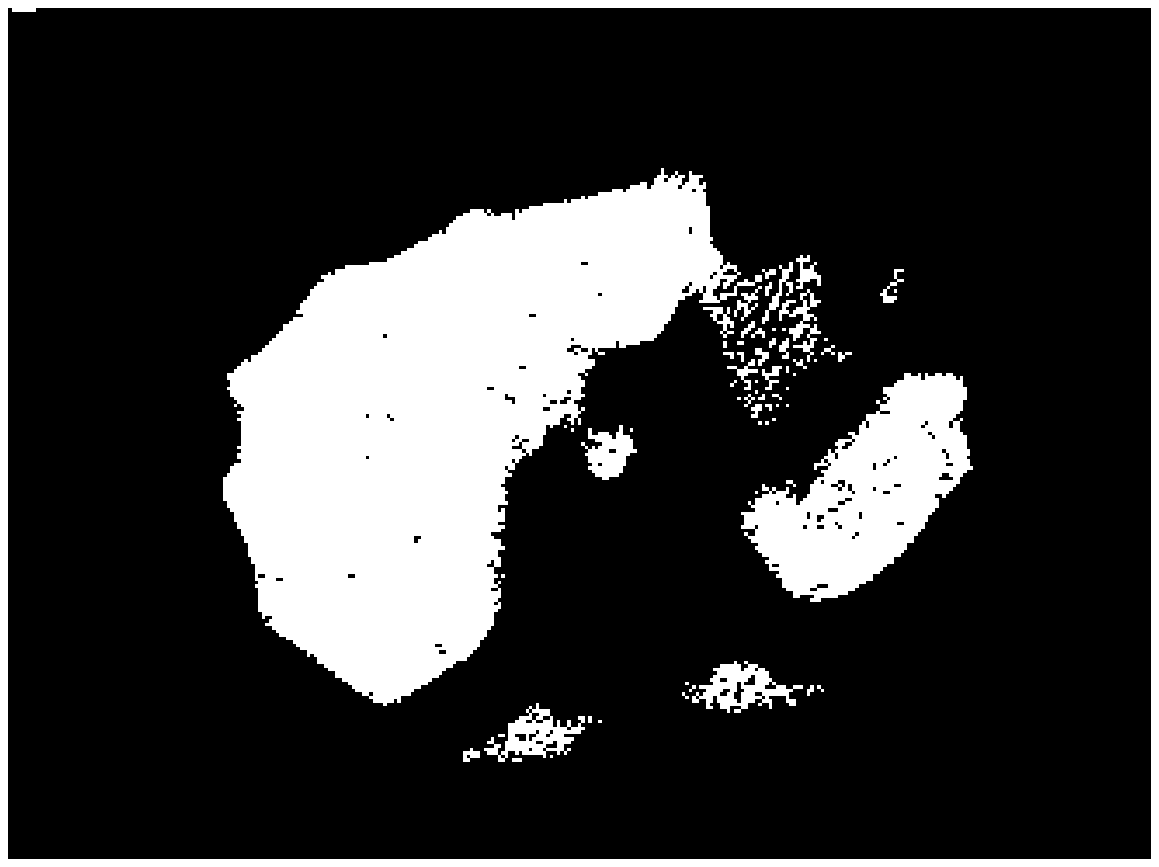}
   \includegraphics[height=2.5 cm]{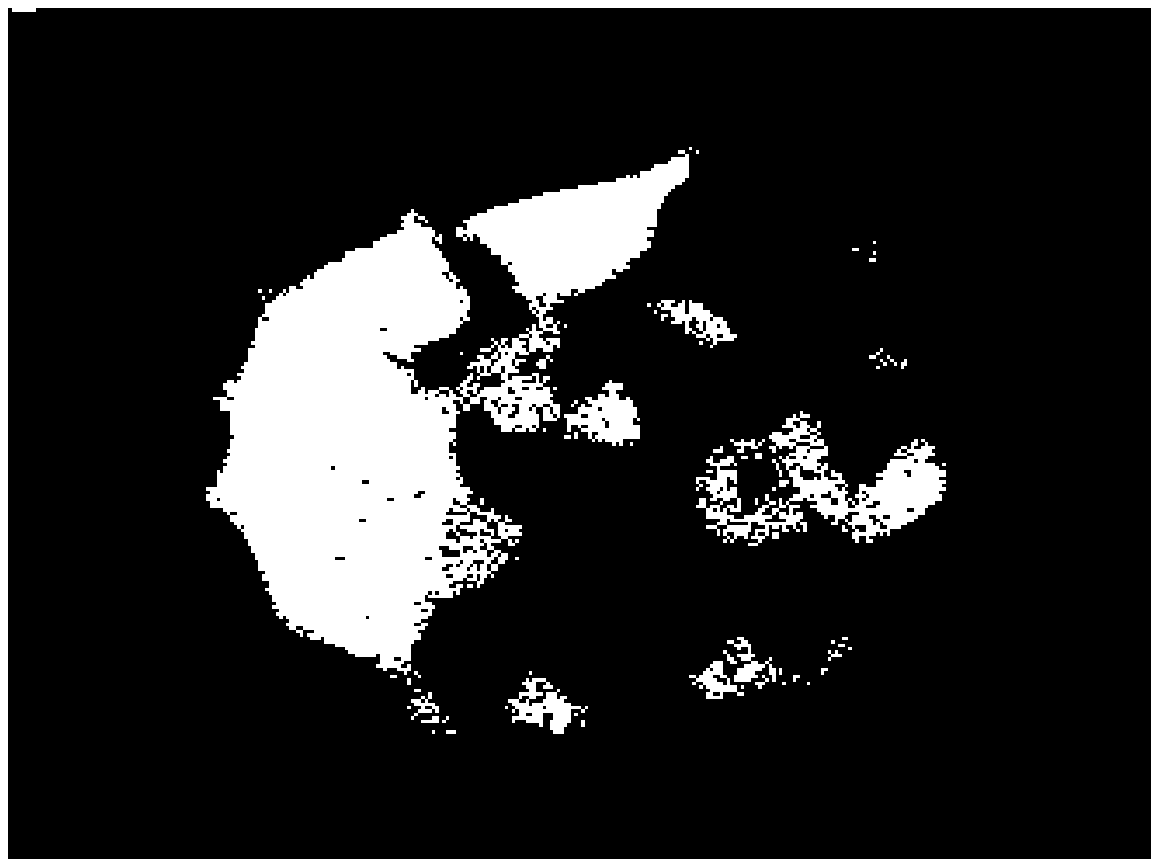}
   \includegraphics[height=2.5 cm]{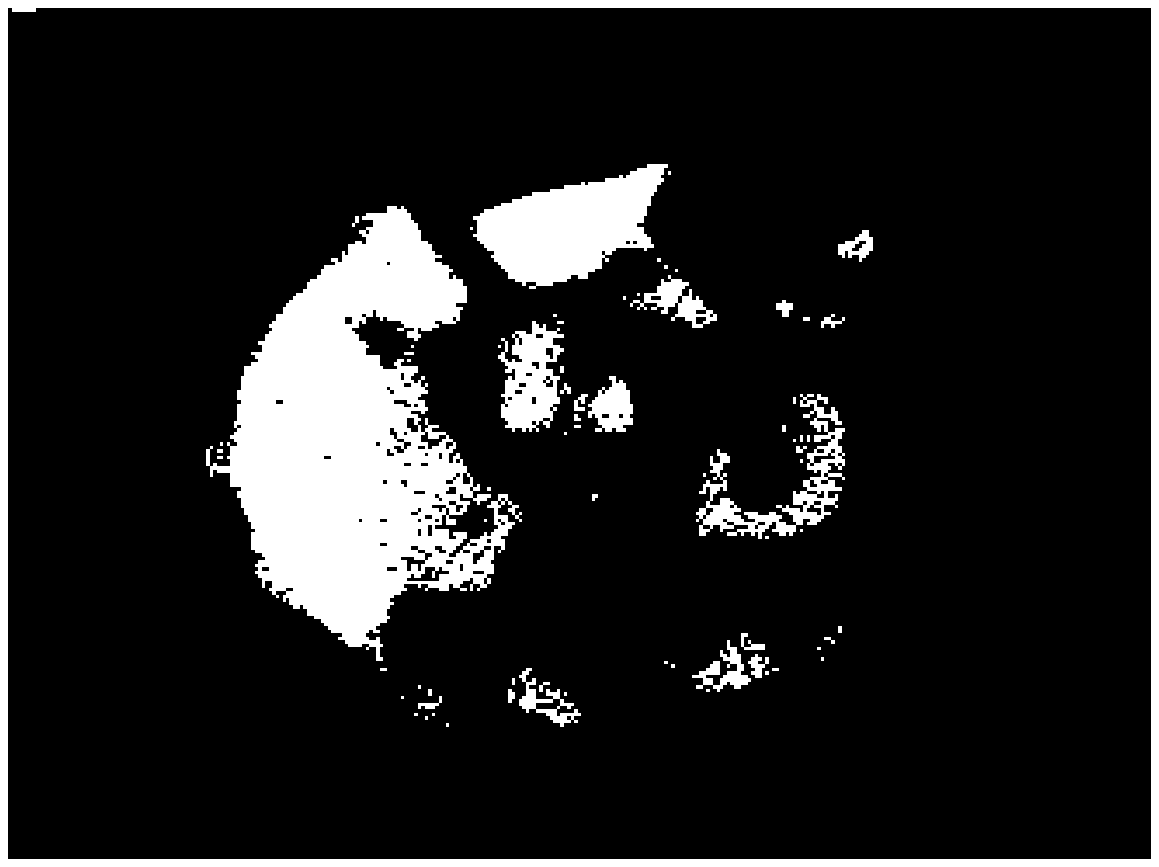}
   \includegraphics[height=2.5 cm]{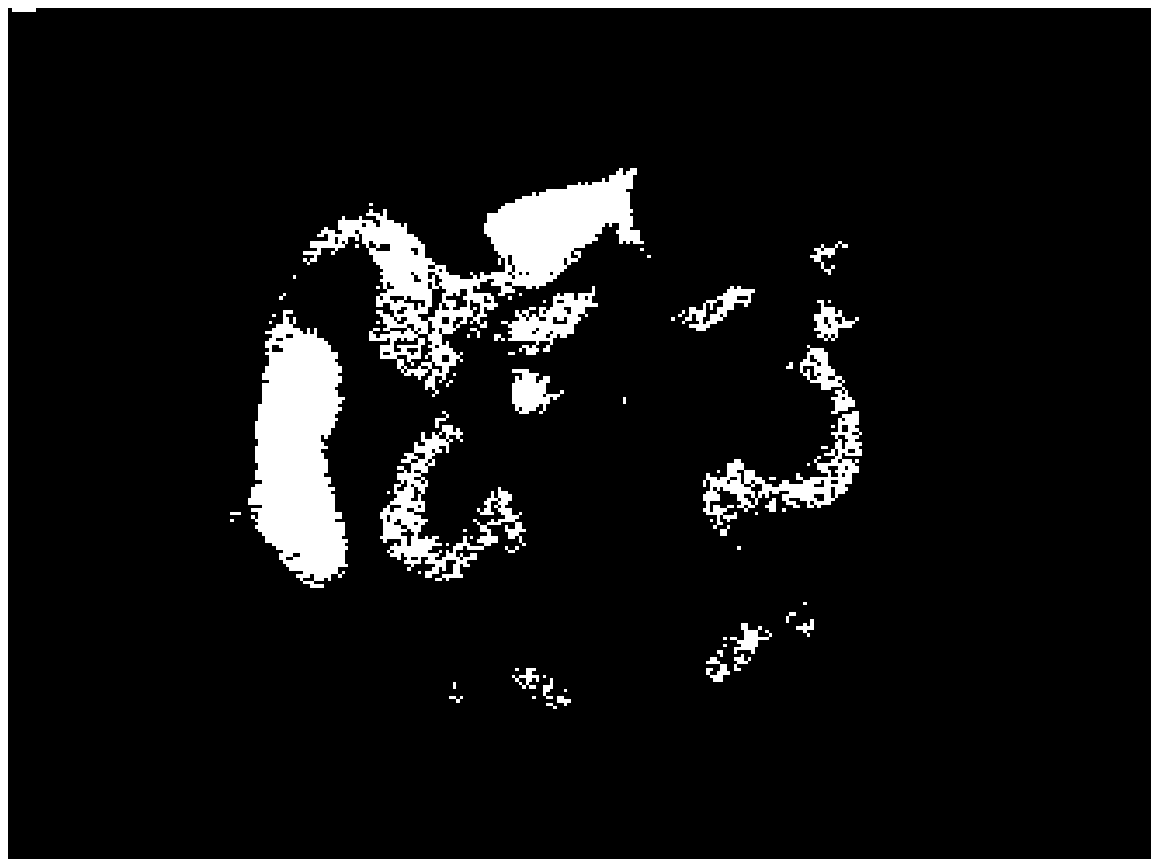}\\ 
      \includegraphics[height=2.5 cm]{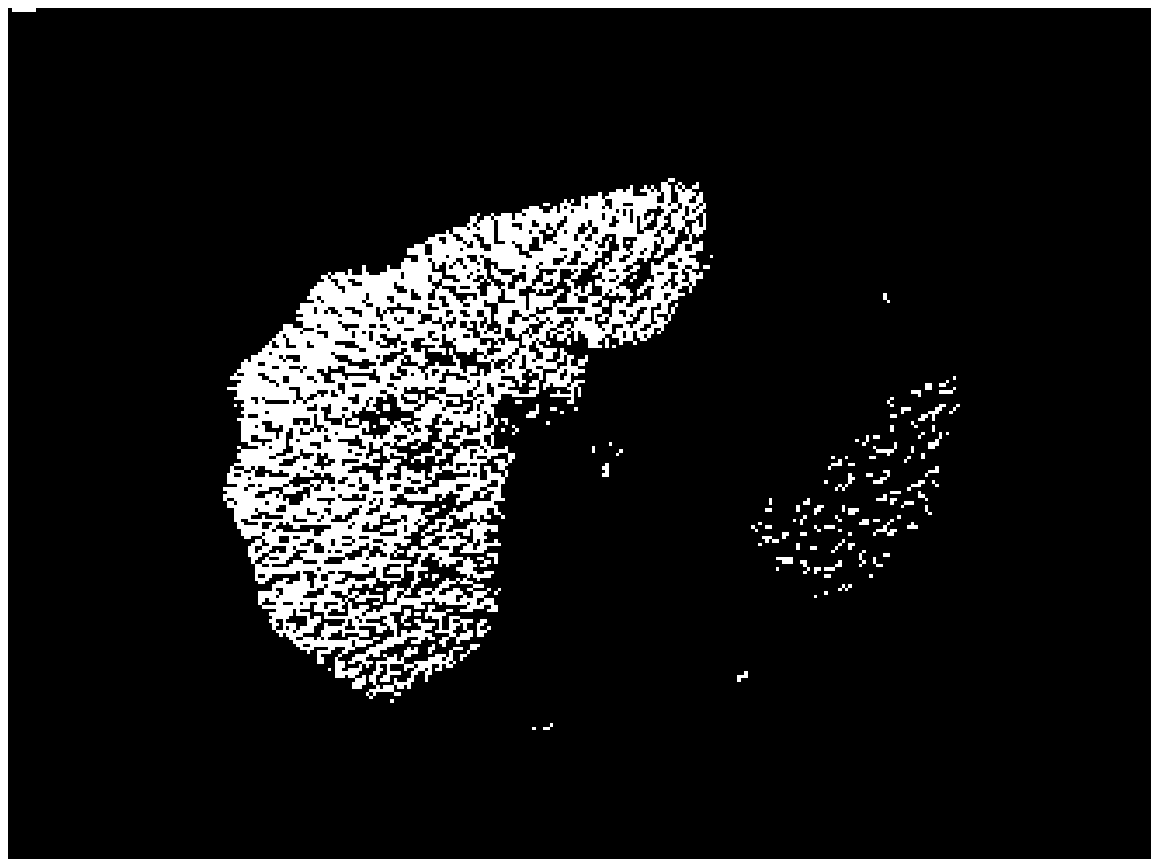}
   \includegraphics[height=2.5 cm]{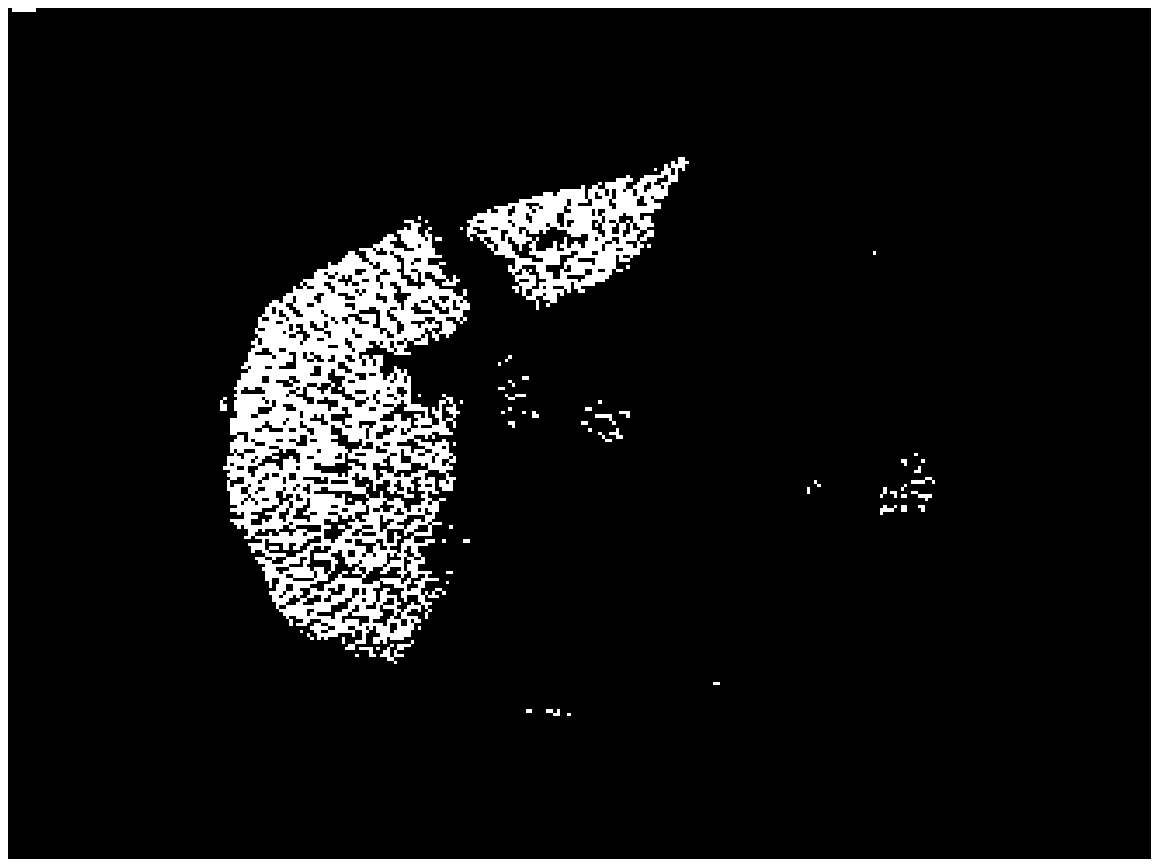}
   \includegraphics[height=2.5 cm]{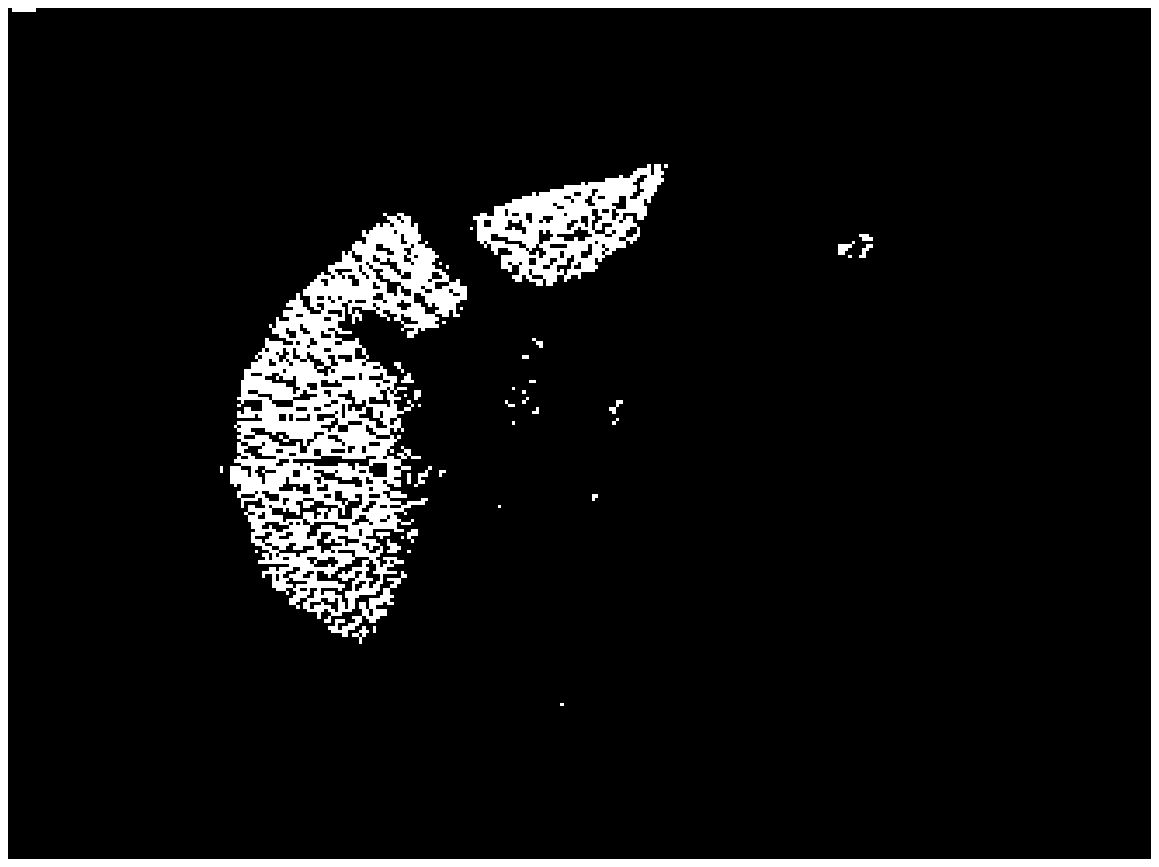}
   \includegraphics[height=2.5 cm]{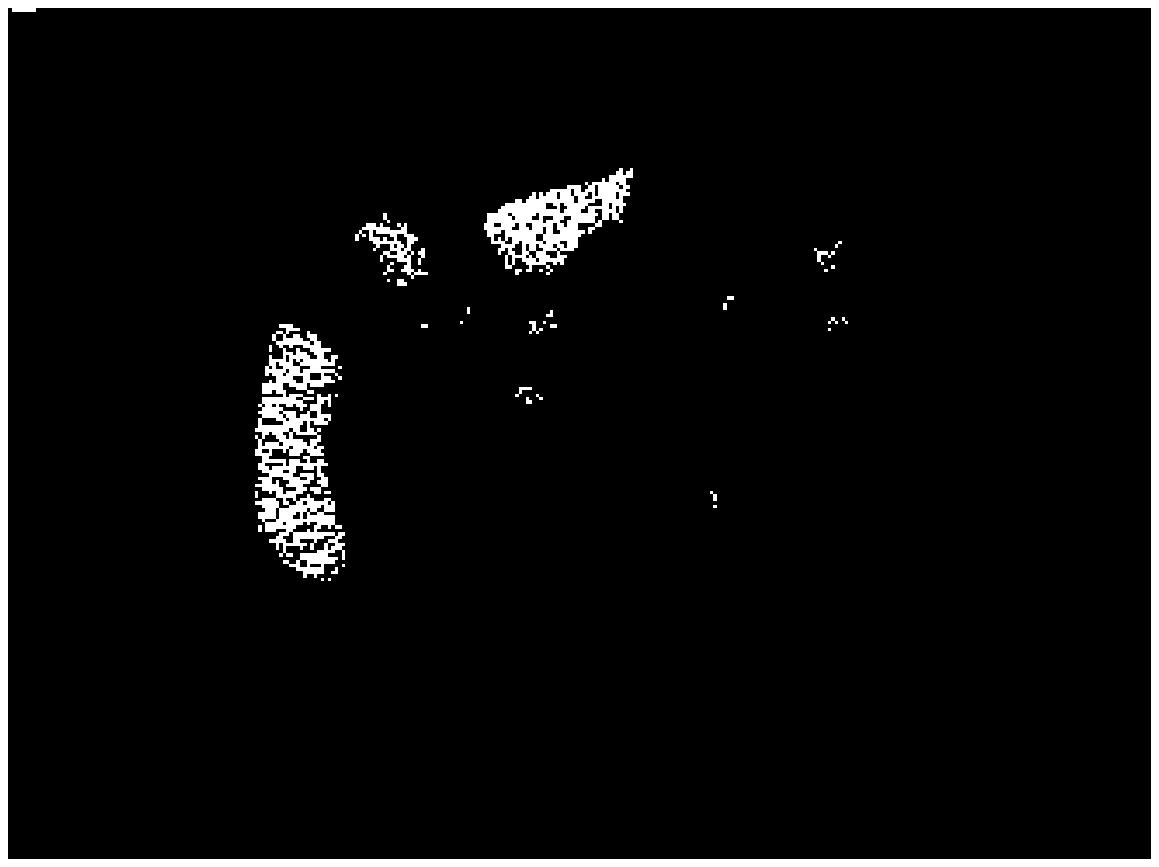}\\ 
      \includegraphics[height=2.5 cm]{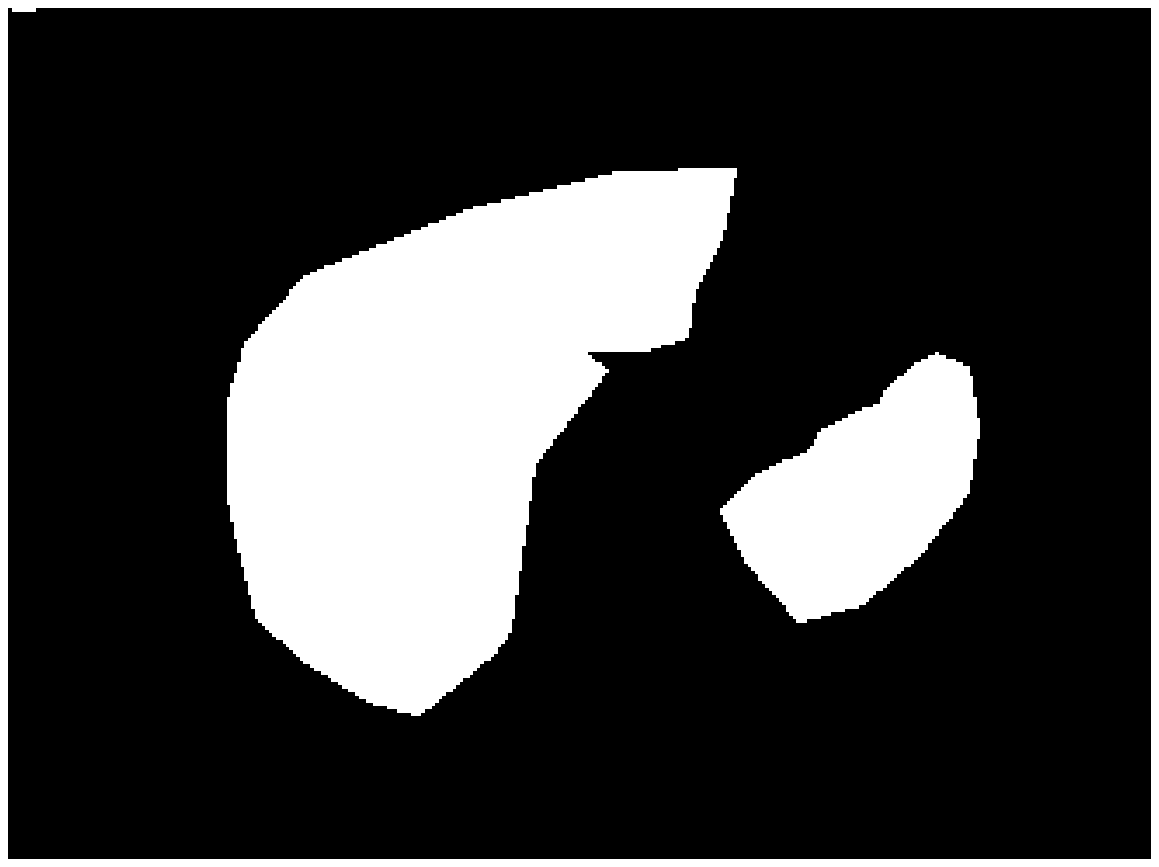}
   \includegraphics[height=2.5 cm]{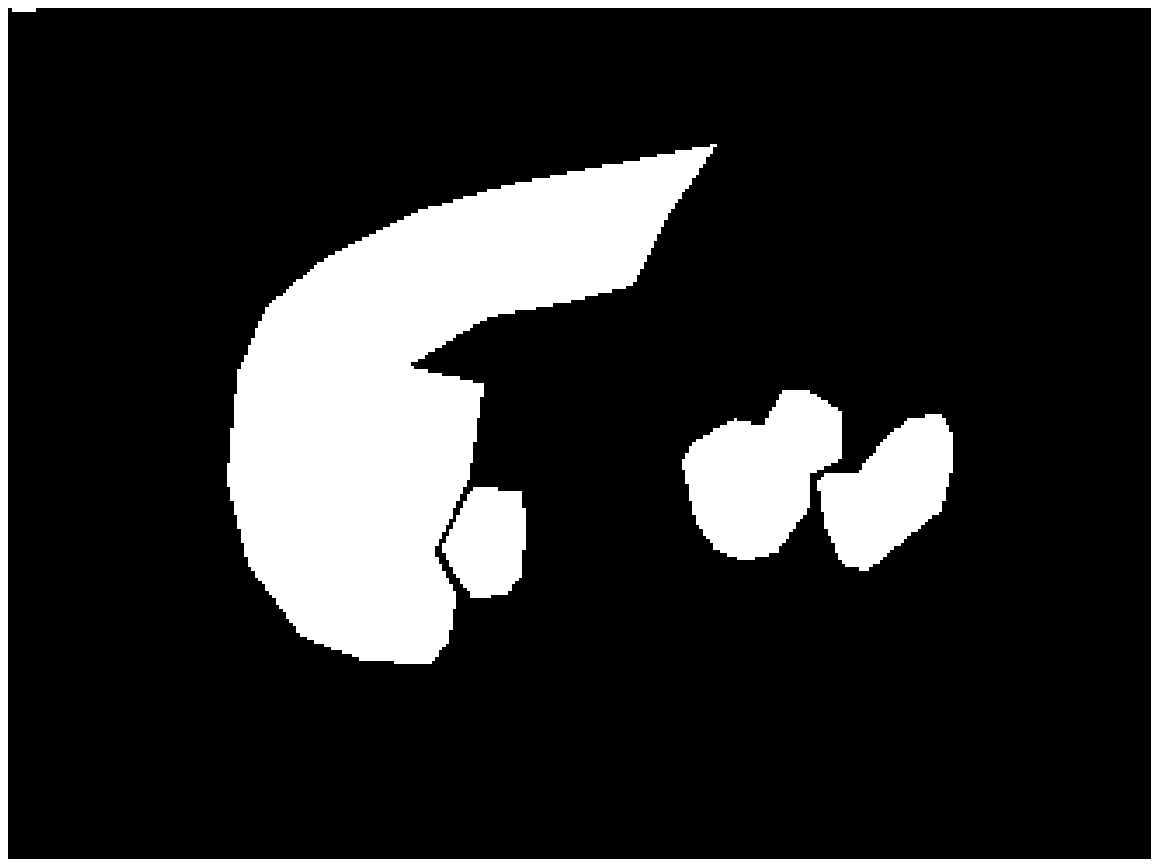}
   \includegraphics[height=2.5 cm]{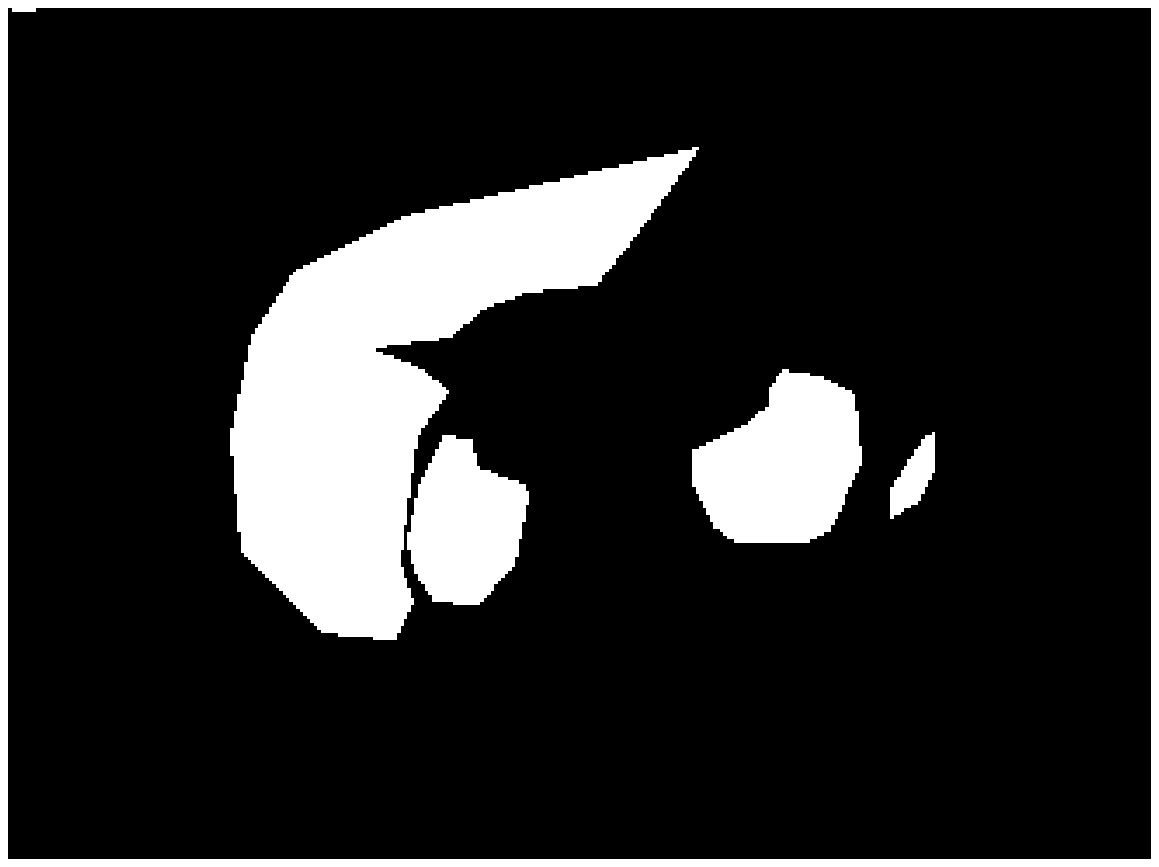}
   \includegraphics[height=2.5 cm]{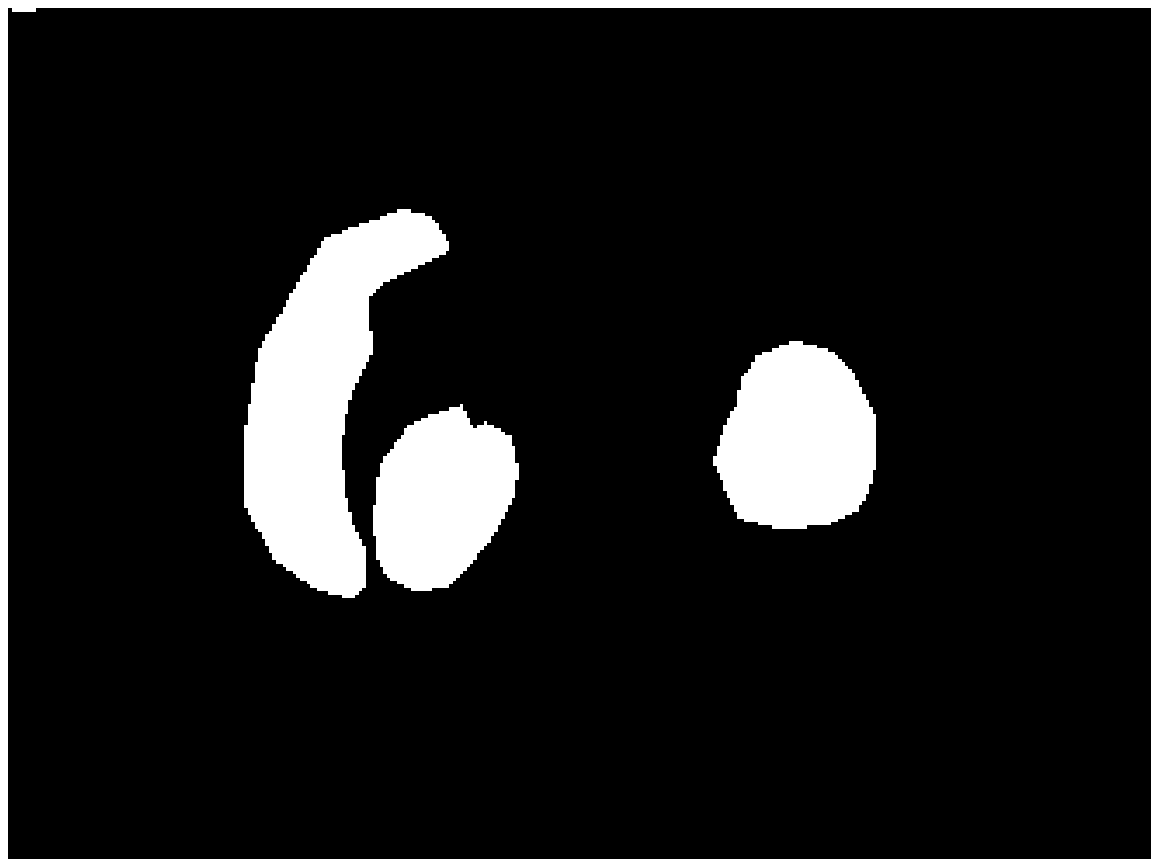}\\ 
   \end{tabular}
   \end{center}
\caption{Different slices of intensity weighted b-scale scenes extracted from a CT image (female subject, abdominal region) are shown in the first row. 2nd-5th rows are showing corresponding thresholded intensity weighted b-scale scenes illustrated in the first row. The last row denotes truly segmented objects of the CT image for the selected scenes, respectively.\label{img:wbs_thrs_abd}} 
\end{figure}

The histogram of $B_{scale}$ image contains only the information about radius of the hyper-balls, hence, it is fairly easy to eliminate small balls and obtain a few largest balls by applying simple \textit{thresholding} technique.  Particularly in this case, thresholding can be used effectively to retain reliable object information. The patterns pertaining to the largest balls retained after thresholding have strong correlations with truly delineated objects shown in the last rows of the Figure~\ref{img:wbs_thrs_abd}. The truly delineated objects and patterns obtained after thresholding share some global similarities; for instance, scale, location, and orientation of the patterns are closely related to truly delineated objects. Patterns show salient characteristics because they depend on object scale estimation and they are mostly spatially localized. Therefore, a concise but reliable relationship can be built using scale, position, and orientation information as parameters. 

Figure~\ref{img:wbs_thrs_abd} demonstrates thresholding process on the intensity weighted b-scale images, $WB_{scale}$. Different slices of $WB_{scale}$ scenes for abdominal CT images are shown in the first rows of the figure. The remaining rows except the last, show thresholded intensity weighted ball-scale scenes, namely $(WB_{scale}^t)$, obtained using various different $t$ intervals. Unlike b-scale scenes, $WB_{scale}$ scenes allow much more flexibility to select thresholding interval $t$, since $t$ is not restricted to be chosen as only object scale in this case. As easily noticed from the fifth row of the Figure~\ref{img:wbs_thrs_abd}, $WB_{scale}^t$ scenes have strong correlations with their truly delineated object correspondences shown in the last row of the same figure. 

In recognition, as the aim is to recognize ``roughly'' the whereabouts of an object of interest in the scene, and also since the trade-off between locality and conciseness of shape variability will be modulated in the delineation step, it will be sufficient to use concise bases produced by PCA without considering localized variability of the shapes. For the former case, on the other hand, it is certain that analyzing variations for each subject separately instead of analyzing variations over averaged ensembles leads to exact solution where specific information present in the particular image is not lost.

\subsection{Relationship Vector}
In order to find the translation, scale, and orientation that best align the shape structure system (learnt from segmented objects in the training set) and the intensity structure system (derived from rough objects obtained through the b-scale encoding scheme from the test set), it is essential to define the relationship between the shape and intensity structures from the training images and incorporate this into the model. We build this relationship via PCA to keep track of translation and orientation differences, and we use ``minimum enclosing box'' approach to find scale component. In minimum enclosing box approach, the real physical units of the truly segmented objects and $WB_{scale}^t$ scenes are used. For orientation analysis, parameters of variations are computed via PCA. Principal components of the shape and intensity structure system, denoted by $\mathbf{PC_o}$ and $\mathbf{PC_b}$, respectively, have an origin and three inertia axes. For the PC systems of the same subject, we have three normalized eigenvectors showing the distribution of variations in each direction, and origins to show centroids of the PC systems.  Briefly, function $\mathbf{F}$ that describes the relationship between $\mathbf{PC_o}$ and $\mathbf{PC_b}$ can be decomposed into the form $\mathbf{F = \left( s, t, R  \right)}$, where $\mathbf{t}: (t_x, t_y, t_z)$ is translation component, $\mathbf{s}$ is a scale component, and~$\mathbf{R}:(R_x, R_y, R_z)$ represents three rotations in $(x,y,z)$. We observe that there is a fixed relationship between $\mathbf{PC_o}$ and $\mathbf{PC_b}$, and the relationship function $\mathbf{F}$ can be split into three sub-functions $\mathbf{f_1, f_2, f_3}$ such that~$\mathbf{F = \left(f_1=s, f_2=t, f_3=R \right)}$.

\subsection{Estimation of Scale Parameter - $\mathbf{f_1=s}$}
Minimum Enclosing Box (MEB)  enclosing the objects of interest for each subject $i=1,\ldots,N$ in the training set is used to estimate the real physical units of the objects in question~\cite{castleman, schalkoff}. The length that connects two farthest corners of the MEB is defined as the scale parameter. 

\subsection{Estimation of Translation Parameters - $\mathbf{f_2=t}:(t_x, t_y, t_z)$}
Estimation of the translation parameters is solely based on forming a linear relationship between the centroids of the binary objects and those obtained from thresholding the intensity weighted b-scale scene. Then, poses of the shapes are computed more accurately due to incorporation of appearance information. 

\subsection{Estimation of Orientation Parameters - $\mathbf{f_3=R}:(R_x, R_y, R_z)$}
Since principal component vectors of shape and intensity structure system constitute an orthonormal basis, and since translation between the $PC$ systems are estimated prior to the estimation of orientation parameters, the two $PC$ systems differ only by orientation, and therefore, the basis vectors in shape structure system can be expressed in terms of the basis vectors in the intensity structure system by the relation
\begin{equation}
\mathbf{PC_b}=\mathbf{(R) (PC_o)},
\end{equation}
where $\mathbf{R}$ is an orthonormal rotation matrix carrying information about the relative positions of shape and intensity structure systems in terms of their Euler angles. A set of $N$ segmented training data sets yields, for each $i=1,\ldots,N$, an orthonormal rotation matrix $\mathbf{R_i}$ that relates ${PC_{o}}_i$ and ${PC_{b}}_i$. We assume that, once the mean rotation and its standard deviation around the mean are found, this information will guide us as to how the $PC_o$ and $PC_b$ of any test image are related. With $F$ modeling the mean relationship between the shape and intensity structures over the training set, the enchanced model assembly becomes $MA=(\mathbb{M}, \Delta F)$, where $\mathbb{M}$ denotes the set of object models and $\Delta F$ denotes the variation observed in $F$ over the training set.

\section{Hierarchical Recognition}
Given any patient test scene $\mathbb{C}$, recognition at the coarsest level proceeds as follows: First the weighted b-scale scene $\mathbb{C}_1$ of $\mathbb{C}$ is computed. Note that this does not require any segmentation. From $\mathbb{C}_1$, the $PC$ system $PC_b$ for the intensity structure is determined. Then, from $F$, the pose of the model assembly $MA$ in $\mathbb{C}$ is determined. Once coarse recognition has been done, the pose of the $MA$ can be refined further by several means: (i) by searching around the pose of coarse recognition but confined to the limits indicated by $\Delta F$; (ii) by segmenting one of the easily segmented objects (such as the skin boundary) and by aligning it with the corresponding object in $MA$. Recognition achieved after all such refinement methods is called fine recognition. Finally, exact refinement gets done in the delineation step which is considered to be the finest level of recognition. This hierarchy can be understood from the perspective of delineation accuracy also such that recognition accuracy itself depends on delineation accuracy, and conversely recognition influences delineation accuracy.

\section{Evaluations and Results}
We used whole body PET-CT scans of 10 female and 10 male patients. The voxel size of the CT images is 1.17 mm x 1.17 mm x 1.17 mm (interpolated from 5 mm slices). We focus on the abdominal region and have selected the following five objects from each subject: skin boundary, liver, left kidney, right kidney and spleen. 

\begin{figure}[htb!]
\begin{center}
   \begin{tabular}{c}
   \includegraphics[height=9 cm]{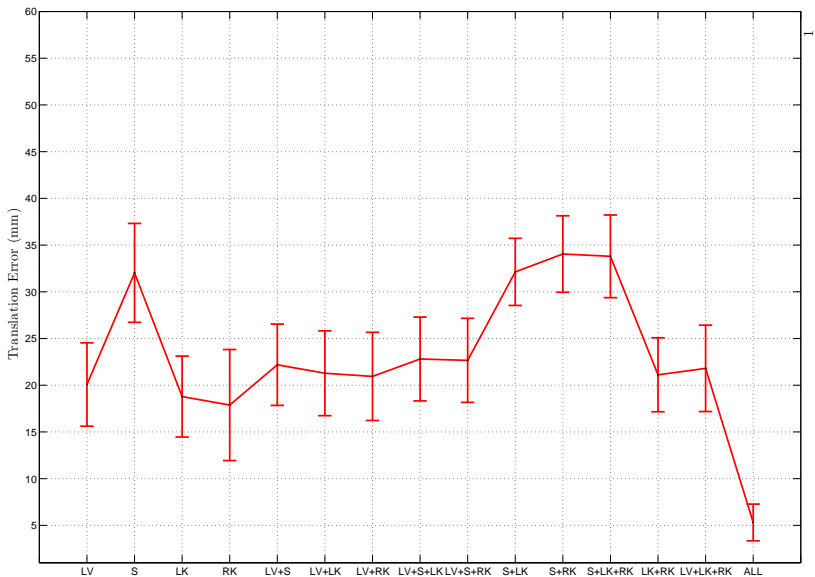}
\end{tabular}
\end{center}
\caption{Recognition accuracy in terms of mean translation error (mm) for CT Abdominal female data with different number and combination of organs included in the model assembly. \label{img:transfemale}} 
\end{figure}

\begin{figure}[htb!]
\begin{center}
   \begin{tabular}{c}
   \includegraphics[height=9 cm]{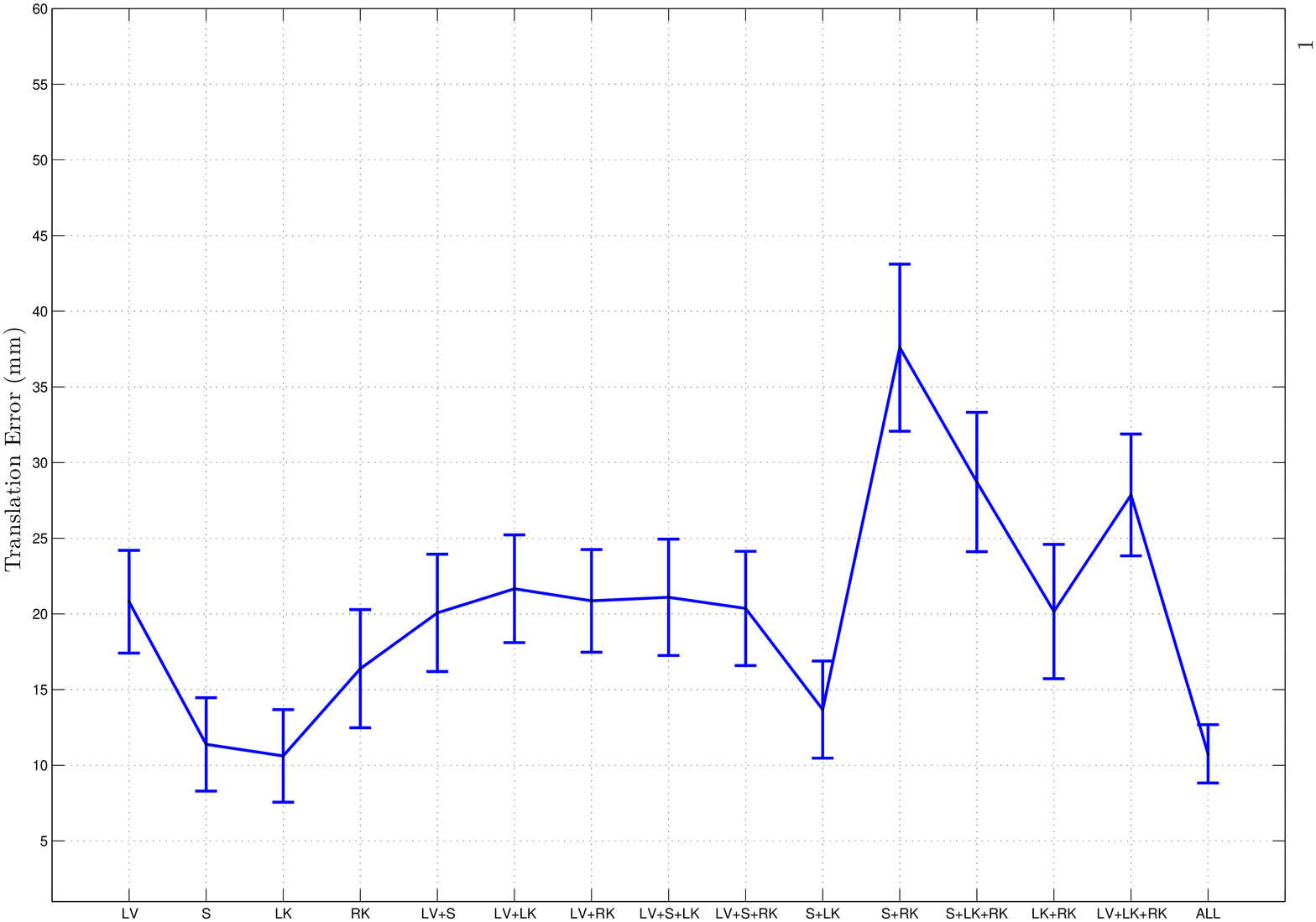}
   \end{tabular}
\end{center}
\caption{Recognition accuracy in terms of mean translation error (mm) for CT Abdominal male data with different number and combination of organs included in the model assembly. \label{img:transmale}} 
\end{figure}

\begin{figure}[htb!]
\begin{center}
   \begin{tabular}{c}
   \includegraphics[height=9 cm]{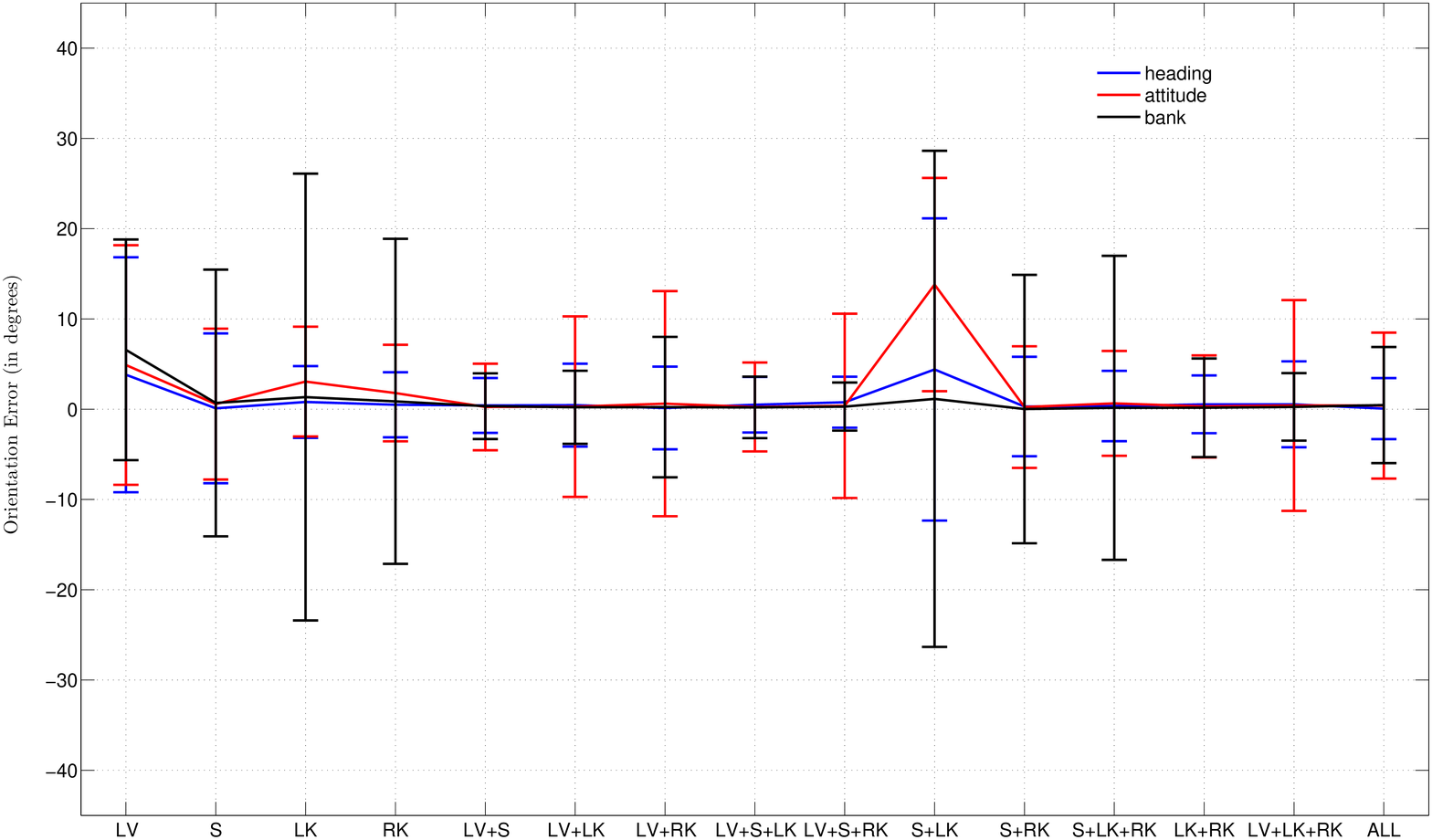}
   \end{tabular}
\end{center}
\caption{Recognition accuracy in terms of mean orientation error (mm) in three directions: heading-x, attitute-y, bank-z for CT Abdominal female data with different number and combination of organs included in the model assembly. \label{img:orientfemale}} 
\end{figure}

\begin{figure}[htb!]
\begin{center}
   \begin{tabular}{c}
   \includegraphics[height=9 cm]{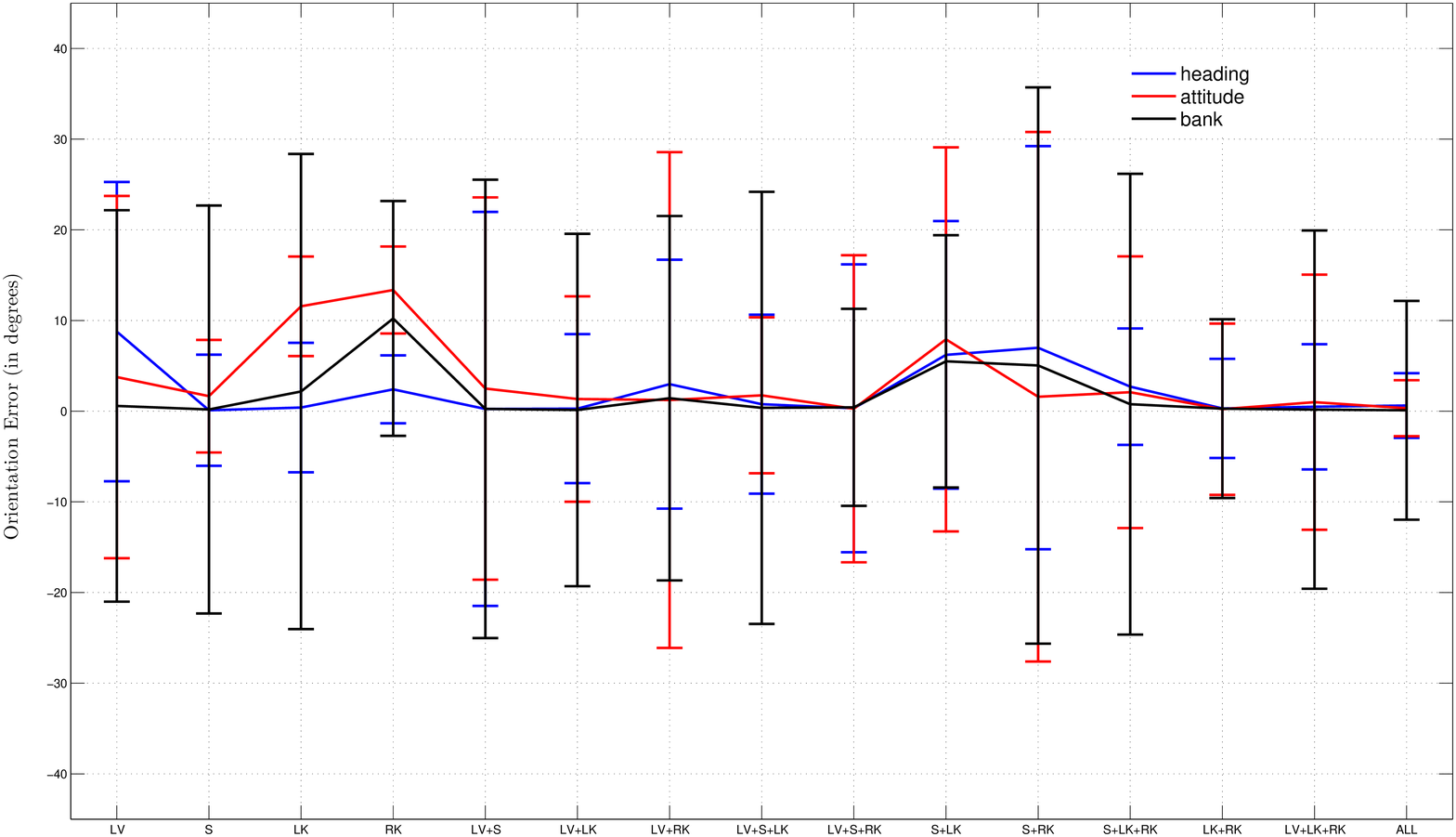}
\end{tabular}
\end{center}
\caption{Recognition accuracy in terms of mean orientation error (mm) in three directions: heading-x, attitute-y, bank-z for CT Abdominal male data with different number and combination of organs included in the model assembly. \label{img:orientmale}} 
\end{figure}

We use leave-one-out-cross-validation (LOOCV) to measure recognition performance for each subject type. Translation, scale, and orientation components of the relationship function $F$ are evaluated separately. The range of scale component in LOOCV tests was found to be  $(0.97-1.07)$. Figures~\ref{img:transfemale} and~\ref{img:transmale} show recognition accuracy in terms of mean translation error over all objects for female and male subjects, respectively, and the different combinations of objects included in the model assembly (shown along the horizontal axis). All results displayed are for coarse recognition only. As easily noticed, the minimum mean translation errors are obtained when all the objects are included in the recognition process. Different combination of objects yields different results. Size and spatial position of the objects play an important role in recognition: it is easier to recognise large objects than smaller objects.  Similarly, Figures~\ref{img:orientfemale} and~\ref{img:orientmale} show recognition accuracy in terms of mean orientation error (in degrees) for female and male data, respectively. Note that the minimum mean orientation error is obtained when all the objects are included in the recognition process.

We observe that the effectiveness of object recognition depends on the number and distribution of objects considered in $MA$. Recognition accuracy is improved with the increasing number of objects. The evaluated results indicate: (1) High recognition accuracy can be achieved by including a large number of objects which are spread out in the body region. (2) Incorporating local object scale information improves the recognition in a way that there is usually no need to do search for scaling, orientation, and translation parameters. (3) The appearance information incorporated via b-scale has strong effect on the computation of PC system, and on the relationship function $F$.  

\section{Conclusion}
(1) The b-scale image of a given image captures object morphometric information without requiring explicit segmentation. b-scales constitute fundamental units of an image in terms of largest homogeneous balls situated at every voxel in the image. The b-scale concept has been previously used in object delineation, filtering and registration. Our results suggest that their ability to capture object geography in conjunction with shape models may be useful in quick and simple yet accurate object recognition strategies. (2) The presented method is general and does not depend on exploiting the peculiar characteristics of the application situation. (3) The specificity of recognition increases dramatically as the number of objects in the model increases. (4) We emphasize that both modeling and testing procedures are carried out on the CT data sets that are part of the clinical PET/CT data routinely acquired in our hospital. The CT data set are thus of relatively poor (spatial and contrast) resolution compared to other CT-alone studies with or without contrast. We expect better performance if higher resolution CT data are employed in modeling or testing.

\section{Acknowledgement}
This paper is published in SPIE Medical Imaging Conference - 2010.

This research is partly funded by the European Commission Fp6 Marie Curie Action Programme (MEST-CT-2005-021170). The second author's research is funded by an NIH grant EB004395.


\end{document}